\documentclass[pmlr]{jmlr}

\RequirePackage{graphicx}
\usepackage{booktabs}
\usepackage{longtable}
\usepackage{amsmath}
\DeclareMathOperator*{\argmin}{arg\,min}
\DeclareMathOperator*{\argmax}{arg\,max}

\makeatletter
\def\set@curr@file#1{\def\@curr@file{#1}} 
\makeatother
\usepackage[load-configurations=version-1]{siunitx} 

\theorembodyfont{\upshape}
\theoremheaderfont{\scshape}
\theorempostheader{:}
\theoremsep{\newline}

\newcommand{\model}{\texttt{bRAGgen}}
\newcommand{\data}{\texttt{bRAGq}}
\definecolor{TableShade}{HTML}{F5F5F5} 
\usepackage{graphicx}
\usepackage{booktabs, multirow} 
\usepackage{soul}
\usepackage{xcolor,colortbl} 
\usepackage{changepage,threeparttable} 
\usepackage{xcolor} 
\usepackage{arydshln}
\usepackage{pgfplots}

\makeatletter
\renewcommand{\ps@jmlrtps}{%
  \renewcommand{\@oddhead}{}%
  \renewcommand{\@evenhead}{}%
  \renewcommand{\@oddfoot}{}%
  \renewcommand{\@evenfoot}{}%
}
\makeatother

\title[Continually Self-Improving Language Models for Bariatric Surgery QA]{Continually Self-Improving Language Models for \\ Bariatric Surgery Question--Answering}

\author{%
  \Name{Yash Kumar Atri} \Email{atri@virginia.edu}\\
  \addr School of Data Science\\
  University of Virginia\\
  Charlottesville, VA, USA
  \AND
  \Name{Thomas Shin} \Email{thomas.shin@uvahealth.org}\\
  \addr Department of Surgery\\
  University of Virginia School of Medicine\\
  Charlottesville, VA, USA
  \AND
  \Name{Thomas Hartvigsen} \Email{hartvigsen@virginia.edu}\\
  \addr School of Data Science\\
  University of Virginia\\
  Charlottesville, VA, USA
}
\begin{document}

\maketitle

\begin{abstract} 



While bariatric and metabolic surgery (MBS) is considered the gold standard treatment for severe and morbid obesity, its therapeutic efficacy hinges upon active and longitudinal engagement with multidisciplinary providers, including surgeons, dietitians/nutritionists, psychologists, and endocrinologists. This engagement spans the entire patient journey, from preoperative preparation to long-term postoperative management. However, this process is often hindered by numerous healthcare disparities, such as logistical and access barriers, which impair easy patient access to timely, evidence-based, clinician-endorsed information. To address these gaps, we introduce \model, a novel adaptive RAG-based model that autonomously integrates real-time medical evidence when response confidence dips below dynamic thresholds. This self‑updating architecture ensures that responses remain current and accurate, reducing the risk of misinformation. Additionally, we introduce \data, a curated dataset of $1,302$ bariatric surgery–related questions, validated by expert bariatric surgeon, constituting the first large‑scale, domain‑specific benchmark for comprehensive MBS care. In a two-phase evaluation, \model\ is benchmarked against state-of-the-art models using both large language model (LLM)-based metrics and expert surgeon review. Across all evaluation dimensions, \model\ demonstrates substantially superior performance in generating clinically accurate and relevant responses. \\Data and Code available at \url{https://github.com/yashkumaratri/bRAGgen}

\end{abstract}

\section{Introduction}

Severe obesity and its subsequent metabolic disease have become a widespread endemic condition, leading to a projected incidence of 25\% across the United States by 2030, causing a massive health burden in the general US population \citep{Ward2019}. Metabolic and bariatric surgery (MBS) remains the gold standard treatment for severe obesity and metabolic disease, with over 270,000 annual procedures in the United States \citep{Clapp2024,Mechanick2020,Barres2013,Loos2022,Setarehdan2023}. 
However, successful weight loss post-MBS heavily relies on patient education, which has led National Bariatric Surgery and Medical societies to emphasize extensive education services for MBS patients \citep{Mechanick2020}. 

Perioperative MBS patient education includes information about dietary modification, adjunctive lifestyle modifications, expectations surrounding postoperative complications, and psychosocial support; all important mitigators of periprocedural complications and postoperative weight regain, which can afflict up to 64\% of patients in 5–10 years post-MBS \citep{Groller2017,Bjerkan2022,David2020,McLennan2023,Kim2023}.
And after surgery, patients only attend their yearly followup an average of 6.5–29.6\% of the time, leaving patients and providers with limited opportunities to communicate.
Overall, the lack of sustained patient engagement and education post-MBS is a critical impediment to optimal postoperative outcomes with many causes, including low health literacy rates, information inaccessibility, and geographic distances to healthcare providers \citep{Setarehdan2023,Mechanick2020,Schlottmann2023,Bartholomay2024}.

Given these challenges, there is a clear need for scalable, accessible, and continually updated educational and decision‑support tools tailored to the unique needs of MBS patients—spanning from preoperative preparation to long-term postoperative management \citep{P2024-sc}. Traditional patient education materials---whether delivered in print, via static websites, or through periodic telehealth visits---often fail to adapt dynamically to emerging clinical evidence
 \citep{Javanparast2021-te} or to the evolving clinical status of individual patients. Moreover, existing digital health platforms seldom incorporate mechanisms to detect when their guidance may be outdated or insufficiently confident \citep{wang2025safetychallengesaimedicine}, leading to knowledge gaps in both patients and clinicians.

Large language models (LLMs) \citep{grattafiori2024llama3herdmodels, abdin2024phi3technicalreporthighly, minaee2025largelanguagemodelssurvey} offer a potential solution by providing natural language interfaces for patients to query.
However, LLMs face limitations due to fixed knowledge cutoffs \citep{cheng2024dateddatatracingknowledge} and the most-capable are trained on broad, general‑purpose corpora \citep{Alber2025-qb}, leaving them unaware of the latest bariatric surgery guidelines or nuanced postoperative considerations \citep{Belisle-Pipon2024-pj}.
One popular way to address outdated knowledge in LLMs is through retrieval‑augmented generation (RAG) methods \citep{gao2024retrievalaugmentedgenerationlargelanguage}, which retrieve up-to-date documents from a database.
However, their databases are typically static and efforts to increase them over time easily introduce “context noise,” overwhelming the LLM with conflicting inputs and producing higher hallucination rates \citep{Zhang2025-wi}. Furthermore, these methods lack built‑in mechanisms to assess when their own outputs may be insufficiently confident, placing patients and clinicians at risk of incomplete or incorrect guidance \citep{lewis2021retrievalaugmentedgenerationknowledgeintensivenlp}.

We propose \model, an adaptive RAG framework that continuously monitors its response confidence and, upon detecting uncertainty, autonomously retrieves and integrates the latest peer-reviewed evidence and clinical guidelines from trusted biomedical sources such as PubMed\footnote{https://pubmed.ncbi.nlm.nih.gov}. This self‑updating architecture ensures that guidance remains current, accurate, and clinically relevant, reducing the risk of outdated or misleading recommendations. Complementing \model, we introduce \data, a curated dataset of $1,302$ bariatric surgery–related questions validated by a bariatric surgeon, constituting the first large‑scale, domain‑specific benchmark for the full spectrum of MBS patient care—from preoperative preparation to long-term postoperative management. In a two-phase evaluation, leveraging both LLM‑based metrics and expert surgeon reviews, \model demonstrates substantially superior performance in generating clinically accurate, relevant, and actionable responses, paving the way for more accessible, evidence‑based support for MBS patients.

\subsection*{Generalizable insights about machine learning in the context of healthcare}
While we focus on bariatric surgery patients' educational needs, our work includes generalizable insights for other healthcare contexts.
First, the need to access up-to-date medical evidence is widespread, especially in areas where the scientific literature grows quickly.
Our work demonstrates that it is feasible to approach this problem by training models to directly access webpages, and keeping them relevant throughout deployment.
Second, our methods are general purpose and are widely applicable.
While there is a major need for bariatric surgery education and it is our expertise, our machine learning methods can be generalized to any domains where patient questions can be collected, relevant literature exists, and model answers can be validated.


\section{Related Work}

The landscape of digital health interventions for bariatric care across the entire surgical spectrum-from preoperative preparation to long-term postoperative management has expanded considerably, with multiple studies evaluating the efficacy and usability of mobile applications and web-based platforms. For instance, a German cohort study by \cite{Wu2024-nk} demonstrated that mHealth follow-up via a dedicated mobile app achieved comparable outcomes to traditional in-person care across weight loss, quality of life, and nutritional status metrics. Similarly, a systematic review by \cite{Patel2020-xo} identified $33$ usability studies of mHealth apps across surgical subspecialties, underscoring both the potential of digital tools and the persistent challenge of sustaining long-term patient engagement.

In parallel, advances in large language models (LLMs) \citep{grattafiori2024llama3herdmodels, abdin2024phi3technicalreporthighly, minaee2025largelanguagemodelssurvey} have led to growing interest in their ability to distill up-to-date information \citep{atri-etal-2023-promoting, dey-etal-2020-corpora, 10261260, 10.1145/3580305.3599830, ATRI2021107152} and support tasks such as (medical) question answering \citep{khlaut-etal-2024-efficient, sviridova-etal-2024-casimedicos, vladika-matthes-2024-improving, saeed-2024-medifact} and clinical decision-making \citep{kim-etal-2024-medexqa, lu-etal-2024-clinicalrag, Singhal2025-bf}. The GPT models \citep{kojima2023largelanguagemodelszeroshot} showcased impressive zero‑ and few‑shot capabilities but is inherently limited by its fixed pretraining cutoff and general‑purpose data \citep{tamkin2021understandingcapabilitieslimitationssocietal}. Retrieval‑augmented generation (RAG) \citep{gao2024retrievalaugmentedgenerationlargelanguage} based methods ground LLM outputs in external document collections to improve factuality \citep{li2024enhancingllmfactualaccuracy, Cai_2024}. However, as the size of the retrieval corpus grows, RAG systems can suffer from “context noise” that increases hallucination rates and lack internal confidence estimates to flag uncertain outputs \citep{wu2024pandorasboxaladdinslamp}.

To overcome the limitations of conventional RAG systems, recent research has explored adaptive retrieval strategies that react to model uncertainty. Iterative and gated retrieval approaches \citep{jiang2025retrievesummarizeplanadvancing, heydari2025contextawarenessgateretrieval} selectively filter external documents and refine the evidence set across multiple rounds, showing improvements in factuality and coherence. While these approaches mitigate irrelevant content and reduce hallucinations, they remain fundamentally external—they treat retrieval as an auxiliary process \citep{10.5555/3495724.3496517} and stop short of modifying the model’s internal knowledge. As such, they lack the capacity to assess and revise the model’s internal parameters in response to evolving clinical evidence. This gap leaves current systems vulnerable to recurring errors when previously seen topics reappear under different linguistic formulations.

In contrast, our proposed framework, \model, introduces an adaptive self-updating mechanism that not only monitors response confidence but actively integrates validated, up-to-date clinical information into the model itself. By embedding uncertainty detection and retrieval within a continual learning loop, \model\ transitions from passive retrieval to active knowledge refinement. This enables it to stay synchronized with the latest postoperative guidelines and avoid repeating outdated or incorrect responses over time.

Complementing this architecture is \data, a rigorously curated dataset of $1,302$ real-world patient questions covering nutrition, lifestyle, complications, and mental health in the postoperative MBS setting. Validated by board-certified bariatric surgeon, \data\ offers the first specialized benchmark to evaluate clinical QA systems beyond general-purpose health datasets. Together, \model\ and \data\ address the dual challenge of knowledge obsolescence and domain specificity—paving the way for clinically grounded, scalable, and responsive patient support in bariatric aftercare.

\begin{table}[!tbp]
\centering
\begin{tabular}{lrr}
\toprule
\textbf{Question Category} & \textbf{No. of Questions} & \textbf{Percentage (\%)} \\
\midrule
Preparation \& Logistics      & 68            & 5.22  \\
Surgical \& Medical Info      & 263           & 20.20 \\
Risks \& Complications        & 221           & 16.97 \\
Recovery \& Lifestyle         & 296           & 22.73 \\
Nutrition \& Diet             & 102           & 7.83  \\
Mental \& Emotional Health    & 287           & 22.04 \\
Cost \& Insurance             & 65            & 4.99  \\
\bottomrule
\end{tabular}
\caption{Distribution of bariatric surgery-related questions across high-level categories in the \data\ dataset. This table presents the number and percentage of questions within each thematic category, highlighting the diverse informational needs of patients throughout the bariatric surgery journey.}
\label{tab:question_distribution}
\vspace{-5mm}
\end{table}

\section{Dataset}

We introduce \data, a domain-specific dataset curated to reflect the breadth and depth of questions commonly posed by bariatric surgery patients. Designed to support the development of intelligent tools for patient education and clinical decision support, \data captures concerns spanning the entire surgical journey—from preoperative preparation to long-term postoperative management. The dataset was constructed in close collaboration with board-certified bariatric surgeons to ensure clinical validity and relevance, encompassing psychological, medical, and lifestyle-related questions. It comprises $1,302$ total entries: $611$ drawn from PubMedQA \citep{jin-etal-2019-pubmedqa}, of which $201$ were flagged by experts as not representative of everyday patient concerns, and $691$ synthetically generated based on expert-informed templates and real-world patient interactions.

The questions span a wide range of thematic categories, ensuring comprehensive coverage of key issues in bariatric care. These include pre-surgical considerations, intraoperative topics, postoperative management, dietary guidance, mental health, and lifestyle adaptation. As shown in Table~\ref{tab:question_distribution}, the largest proportion of questions fall under \textit{Recovery \& Lifestyle} (22.73\%), followed by \textit{Mental \& Emotional Health} (22.04\%) and \textit{Surgical \& Medical Info} (20.20\%), reflecting the areas patients most frequently seek guidance on. Table~\ref{tab:sample_questions} presents representative examples from each category, illustrating the dataset’s granularity and diversity. By aligning with real-world patient priorities and clinical input, \data provides a rigorous benchmark for evaluating the performance of language models in delivering accurate, trustworthy, and context-aware responses in the bariatric surgery domain. Beyond benchmarking, it also serves as a valuable resource for training patient-facing conversational agents that are empathetic, evidence-informed, and sensitive to the unique needs of this clinical population.

\begin{table}[!ht]
\centering
\footnotesize
\begin{tabular}{p{0.22\textwidth} p{0.72\textwidth}}
\toprule
\textbf{Question Category} & \textbf{Sample Questions} \\
\midrule
Risks \& Complications      & 1. Are vitamin D levels and bone turnover markers related to non-alcoholic fatty liver disease 
                               in severely obese patients? \\
                             & 2. Does older age limit postbariatric surgery cognitive benefits: a preliminary investigation? \\
\hdashline
Recovery \& Lifestyle       & 1. Does clinical trial demonstrate exercise following bariatric surgery improves insulin 
                               sensitivity? \\
                             & 2. Are serum markers of bone turnover increased at six and 18 months after Roux-en-Y bariatric 
                               surgery: correlation with the reduction in leptin? \\
\hdashline
Preparation \& Logistics    & 1. Does a Pre-Hospital Patient Education Program improve Outcomes of Bariatric Surgery? \\
                             & 2. Does perioperative care map improve compliance with best practices for the morbidly obese? \\
\hdashline
Surgical \& Medical Info    & 1. Is laparoscopic gastric bypass superior to laparoscopic gastric banding for treatment of 
                               morbid obesity? \\
                             & 2. Is potentially life-threatening sleep apnea unrecognized without aggressive evaluation? \\
\hdashline
Cost \& Insurance           & 1. Does medicare and Medicaid status predict prolonged length of stay after bariatric surgery? \\
                             & 2. Is medication cost significantly reduced after Roux-en-Y gastric bypass in obese patients? \\
\hdashline
Mental \& Emotional Health  & 1. Are patient expectations of bariatric surgery gender specific -- a prospective, multicenter 
                               cohort study? \\
                             & 2. Is support group meeting attendance associated with better weight loss? \\
\hdashline
Nutrition \& Diet           & 1. Does dehydroepiandrosterone-sulfate modify human fatty acid composition of different adipose 
                               tissue depots? \\
                             & 2. Does low 25-hydroxyvitamin D affect insulin sensitivity in obesity after bariatric surgery? \\
\bottomrule
\end{tabular}
\caption{Sample Questions for Each Bariatric Surgery Category. This table presents two example questions from each major category within the bariatric surgery domain, reflecting the primary concerns of patients throughout their surgical journey.}
\label{tab:sample_questions}
\end{table}

\begin{figure}[t]
    \centering
    \includegraphics[width=0.95\textwidth, trim=20 105 100 0, clip]{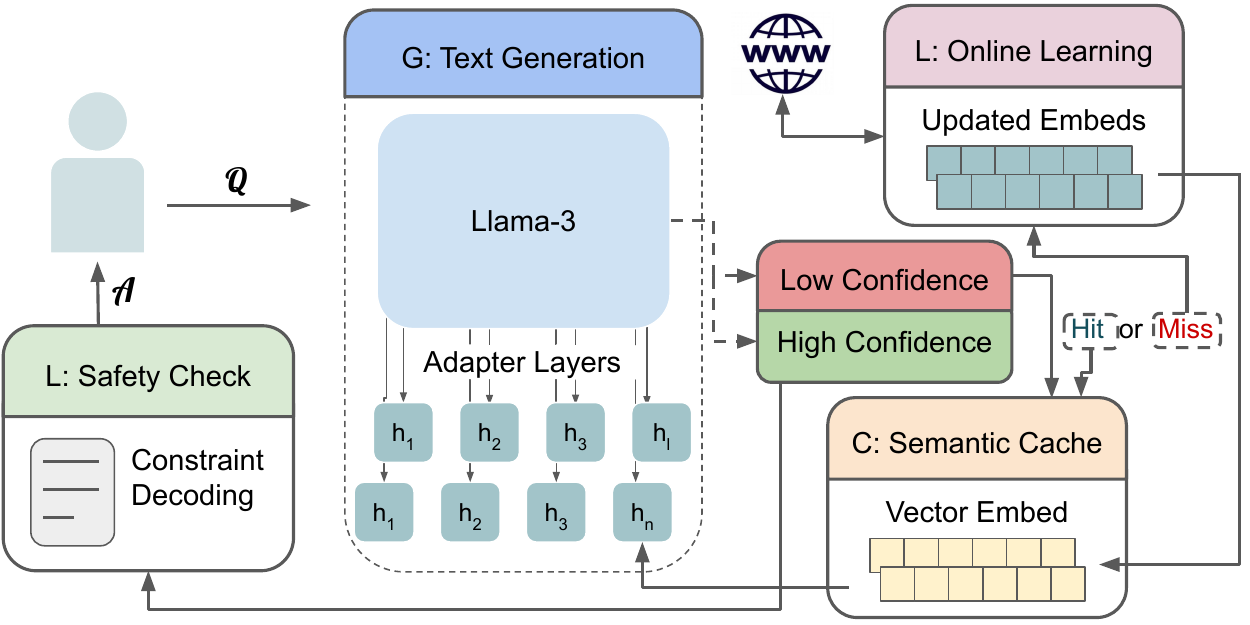}
    \caption{Architecture of the proposed method \model, The system integrates large language models (eg. Llama3) with real-time web retrieval capabilities. When confidence falls below the threshold ($\alpha$), the system automatically retrieves updated information from authoritative medical sources to enhance response accuracy.}
    \label{fig:system_architecture}
\end{figure}

\section{Proposed Methodology}
In response to the growing need for timely, evidence-based decision support in clinical settings, we propose an integrated framework that enhances retrieval-augmented generation (RAG) with continuous online adaptation. Our system is specifically designed to provide contextually relevant, accurate, and safe clinical recommendations by combining several key components: a semantic cache, a multi-source web retrieval engine, an adaptive text generation module, and an online learning protocol. These components are carefully chosen to address critical challenges in clinical decision-making, including rapid access to high-quality medical information, comprehensive evidence retrieval from trusted sources, dynamic and context-sensitive response generation, and continuous model refinement.

We formalize the framework as:
\[
\mathcal{S} = (\mathcal{C}, \mathcal{R}, \mathcal{G}, \mathcal{L})
\]
where each component is defined as follows:
\begin{itemize}
    \item $\mathcal{C}$ represents the semantic cache, which ensures fast retrieval of relevant medical documents. It leverages SentenceTransformer embeddings and Faiss indexing for rapid access to domain-specific information, minimizing latency.
    \item $\mathcal{R}$ is the web-based multi-source retrieval engine, which uses a Markov Decision Process (MDP) to focus on authoritative medical sources. The engine aggregates external evidence from trusted web sources, enriching the context when the cache does not suffice.
    \item $\mathcal{G}$ is the adaptive text generation module, which employs low-rank adaptation (LoRA) techniques to fine-tune a large pre-trained language model for domain-specific tasks. This module tailors the generated responses to clinical contexts while maintaining accuracy and compliance with domain constraints.
    \item $\mathcal{L}$ is the online learning module, which continuously refines the model. It updates the system with new data and interactions, enabling the framework to improve over time and adapt to evolving clinical guidelines.
\end{itemize}

By combining these components, our framework offers a dynamic, evidence-based decision support system that remains adaptable and effective over time. Each module addresses specific challenges, ensuring accurate, context-aware, and clinically safe recommendations in real-time clinical environments.

\subsection{Semantic Knowledge Caching}
To minimize response latency and ensure the rapid retrieval of high-quality clinical evidence, our system incorporates a semantic knowledge caching mechanism. This cache maintains a collection of document-query pairs, denoted as $\mathcal{D} = \{(q_i, d_i)\}_{i=1}^N$, where each query $q_i \in \mathbb{R}^{768}$ is generated using the BioClinicalBERT model and is paired with its corresponding clinical document $d_i$. Given an input query $\mathbf{q}$, the cache efficiently retrieves the document $d_j$ that maximizes the cosine similarity between the query and document embeddings:

\begin{equation}
    \mathcal{C}(\mathbf{q}) = \argmax_{d_j \in \mathcal{D}} \frac{\mathbf{q} \cdot \mathbf{q}_j}{\|\mathbf{q}\| \|\mathbf{q}_j\|} \quad \text{subject to} \quad \frac{\mathbf{q} \cdot \mathbf{q}_j}{\|\mathbf{q}\| \|\mathbf{q}_j\|} \geq \tau_c,
\end{equation}
where $\tau_c = 0.7$ is a cosine similarity threshold that ensures only the most relevant documents are retrieved.

To maintain the cache’s relevance, new query-document pairs $(\mathbf{q}, d)$ are continuously added. Obsolete entries are removed based on an eviction policy designed to prioritize documents that remain valuable over time. Specifically, the cache is updated as follows:

\begin{equation}
    \mathcal{D} \leftarrow \mathcal{D} \cup \{(\mathbf{q}, d)\} \setminus \{\argmin_{d_k} \psi(d_k)\},
\end{equation}
where the eviction score $\psi(d_k)$ for a document $d_k$ is defined as:

\[
\psi(d_k) = \alpha f_u(d_k) + (1-\alpha) e^{-t/\beta},
\]
with $\alpha=0.6$, $f_u(d_k)$ representing the document's usage frequency, and $e^{-t/\beta}$ accounting for the document's age, where $t$ is the time since the last access, and $\beta$ is document total stored time. 

This eviction policy ensures that frequently accessed and recent documents are retained in the cache, while less relevant or outdated documents are pruned. 

The caching mechanism is implemented using SentenceTransformer embeddings, with Faiss indexing, enabling efficient similarity search. The cache is constrained by a fixed size (e.g., 500 documents), ensuring fast retrieval and minimal computational overhead. By maintaining a small, high-quality set of relevant documents, the semantic cache significantly reduces response times during clinical decision support, providing timely access to critical information.

\subsection{Multi-Source Web Retrieval Engine}
When the semantic cache does not provide sufficient information to answer a query, the multi-source web retrieval engine is triggered to gather additional evidence from trusted external sources. This component is designed to ensure that the system can access comprehensive, up-to-date information from diverse domains, including those not covered by the cached documents. The retrieval process is framed as a Markov Decision Process (MDP), which enables focused and dynamic crawling across medical websites to retrieve relevant content. 

At each time step $t$, the action $a_t$ taken by the retrieval engine is determined by maximizing the expected cumulative reward, which is computed as follows:

\begin{equation}
    a_t = \argmax_{a \in \mathcal{A}} \sum_{s'} P(s' \mid s, a) \left[ R(s, a) + \gamma V(s') \right],
\end{equation}
where the reward function $R(s, a)$ is defined as:

\begin{equation}
    R(s, a) = \mathbb{I}_{\text{TLD}(s) \in \{\text{.gov}, \text{.edu}\}} \cdot \text{BM25}(s, q),
\end{equation}
where $\mathbb{I}_{\text{TLD}(s) \in \{\text{.gov}, \text{.edu}\}}$ is an indicator function that ensures the retrieved documents are from authoritative sources (i.e., websites with `.gov` or `.edu` top-level domains), while $\text{BM25}(s, q)$ is a content relevance score computed using the BM25 ranking function to assess how well the document $s$ answers the query $q$.

The retrieval process is powered by the DuckDuckGo API, which allows for broad web searches while prioritizing authoritative sources through URL domain filtering. The BM25 scoring system is applied to rank the retrieved documents based on their relevance to the input query, ensuring that the most pertinent and reliable documents are selected. Once the relevant documents are retrieved, they are incorporated into the semantic cache, thereby enriching the context available for generating the system’s response.

This mechanism enhances the system’s ability to provide informed, evidence-based recommendations, particularly in scenarios where the cached knowledge does not suffice, by tapping into the vast amount of publicly available, authoritative medical content across the web.

\subsection{Adaptive Text Generation}
To generate accurate and contextually relevant clinical recommendations, we employ the Llama3-8B model, enhanced using low-rank adaptation (LoRA). This approach is designed to efficiently fine-tune a large pre-trained language model to domain-specific tasks, while minimizing computational overhead and memory usage. The adaptation is performed in a low-rank fashion, enabling the model to adjust quickly to specific medical domains without the need for full retraining.

For each transformer layer $l$ in the model, the adaptive hidden representation is computed as follows:

\begin{equation}
    h_l^{\text{adapt}} = h_l^{\text{base}} + \Delta W_l \, h_l^{\text{base}}, \quad \Delta W_l = B_l A_l,
\end{equation}

where $B_l \in \mathbb{R}^{d \times r}$ and $A_l \in \mathbb{R}^{r \times d}$ are the learned low-rank matrices, and $r = 32$ is the rank used for adaptation. The matrices $B_l$ and $A_l$ capture domain-specific information while ensuring that the adaptation process remains efficient and scalable. The term $h_l^{\text{base}}$ represents the original, pre-trained hidden representation of the model at layer $l$.

To evaluate the quality of the generated response, we utilize a perplexity measure, which quantifies the uncertainty in predicting the next token in the sequence:

\begin{equation}
    \mathcal{P}(\mathbf{r} \mid q) = \exp\left( -\frac{1}{T} \sum_{t=1}^{T} \log p_\theta(r_t \mid r_{<t}, q) \right),
\end{equation}

where $\mathcal{P}(\mathbf{r} \mid q)$ is the perplexity of the generated response $\mathbf{r}$ given the input query $q$, and $T$ is the length of the response. The term $p_\theta(r_t \mid r_{<t}, q)$ represents the model's predicted probability of the token $r_t$ at position $t$, conditioned on the preceding tokens and the query.

If the perplexity of the response exceeds a threshold $\tau_p = 4.5$, it indicates that the model's output is not sufficiently confident or relevant. In such cases, the system triggers additional retrieval and adaptation cycles to refine the response, improving its accuracy and relevance by incorporating more domain-specific knowledge.

This adaptive approach ensures that the system can generate high-quality clinical recommendations that are both contextually appropriate and tailored to the specific needs of the patient or healthcare provider.

\subsection{Online Learning Protocol}
To ensure that the model remains up-to-date and adaptable in the face of new evidence and user interactions, we implement an online learning module. This protocol allows the model to continuously refine its performance by integrating fresh data and adjusting its parameters over time. The training objective is designed to balance the model's ability to predict accurate outcomes while avoiding overfitting to recent data, using a regularized cross-entropy loss function:

\begin{equation}
    \mathcal{L}_{\text{adapt}} = \mathbb{E}_{(q,d) \sim \mathcal{B}} \left[-\log p_\theta(d \mid q)\right] + \lambda \|\Theta_A\|_F^2,
\end{equation}

where $\mathbb{E}_{(q,d) \sim \mathcal{B}}$ represents the expectation over a mini-batch $\mathcal{B}$ of query-document pairs, $p_\theta(d \mid q)$ is the predicted probability of document $d$ given query $q$, and $\lambda$ is the regularization parameter that controls the magnitude of the model's parameters. The term $\|\Theta_A\|_F^2$ represents the Frobenius norm of the model’s adaptation parameters $\Theta_A$, which serves as a regularizer to prevent overfitting during updates.

The experience buffer $\mathcal{B}$ is updated dynamically to maintain a diverse and representative sample of query-document pairs. This buffer is managed using a Faiss-based nearest neighbor search mechanism, which ensures that new samples are included in a way that preserves diversity and reduces redundancy. The update rule is as follows:

\begin{equation}
    \mathcal{B} \leftarrow \mathcal{B} \cup \{(q_i,d_i)\} \setminus \{\argmax_{(q_j,d_j)} \text{sim}(q_j,q_i)\},
\end{equation}

where $\text{sim}(q_j,q_i)$ denotes the similarity between queries $q_j$ and $q_i$, and we remove the pair that is most similar to the newly added sample, ensuring the buffer contains varied and non-redundant training data. This approach helps the model avoid memorizing specific query-answer pairs and encourages generalization across a broad range of contexts.

The online learning module ensures that the system adapts in real-time to emerging evidence, evolving patient needs, and new clinical knowledge. As a result, the model continually improves its performance, staying current with the latest developments and capable of providing up-to-date, accurate recommendations.

\subsection{Safety and Response Validation}
Ensuring that generated outputs are both safe and clinically valid is of paramount importance in our system. To achieve this, we apply constrained decoding during the text generation process, which ensures that generated responses adhere to safety guidelines and domain-specific constraints. The constrained decoding objective is formulated as follows:

\begin{equation}
    \mathbf{r}_{\text{safe}} = \argmax_{\mathbf{r} \in \mathcal{V}^*} p_\theta(\mathbf{r} \mid q) \prod_{i=1}^{n} \phi_i(r_i),
\end{equation}

where $\mathbf{r}$ is the generated response, and $\mathcal{V}^*$ is the vocabulary space. Each constraint function $\phi_i(r_i)$ is designed to enforce specific safety requirements on individual tokens $r_i$ in the response. The constraint function $\phi_i$ is defined as:

\begin{equation}
    \phi_i(r_i) = \mathbb{I}\Bigl\{ \nexists\, w \in r_i : w \in \mathcal{W}_{\text{speculative}} \Bigr\},
\end{equation}

where $\mathbb{I}$ is the indicator function, and $\mathcal{W}_{\text{speculative}}$ is a set of words or phrases that are deemed speculative or unsafe in a clinical context. This constraint ensures that the generated response does not include any terms or statements that might mislead patients or suggest unverified clinical practices.

To ensure clinical validity, we further compare outputs with reference texts using BERTScore \citep{zhang2020bertscoreevaluatingtextgeneration}, which evaluates semantic similarity via contextual embeddings. Together, constrained decoding and BERTScore validation ensure that responses remain both safe and aligned with evidence-based clinical content.


\section{Baselines}
\label{sec:baselines}

We evaluated several baseline models to assess their performance on our medical question-answering benchmark:

\noindent (i) \textbf{RAG$^2$} \citep{sohn2024rationaleguidedretrievalaugmentedgeneration} relies on a pre-cached offline corpus of biomedical documents for retrieval, avoiding real-time web queries. It enhances standard RAG methods by using perplexity-based labels and LLM-generated rationales to selectively retrieve and filter context, improving relevance and reducing noise. (ii) \textbf{MedGraphRAG} \citep{wu2024medicalgraphragsafe} also uses an offline cache of biomedical documents for retrieval. It further integrates a structured medical knowledge graph to guide the retrieval process, leveraging clinical relationships to improve the contextual relevance and factual accuracy of the retrieved information. \noindent (iii) \textbf{Llama3-8B} \citep{grattafiori2024llama3herdmodels} is a 8-billion-parameter large language model evaluated under two configurations: (a) \textbf{Zero-shot}, where the model responds using only its pre-trained knowledge; and (b) \textbf{Context-prompted}, where external context retrieved from offline sources is appended to the prompt to improve answer quality. \noindent (iv) \textbf{Phi-3} \citep{abdin2024phi3technicalreporthighly} is a lightweight 3.8-billion-parameter model optimized for efficiency and edge deployment. We evaluate Phi-3 in both (a) \textbf{Zero-shot} and (b) \textbf{Context-prompted} modes, assessing its ability to handle medical queries with and without retrieval-based augmentation. \noindent (v) \textbf{Mistral Instruct} \citep{jiang2023mistral7b} is a 7-billion-parameter instruction-tuned model designed for strong performance on alignment-focused tasks. It is tested in (a) \textbf{Zero-shot} mode, where it relies solely on instruction tuning, and (b) \textbf{Context-prompted} mode, where it incorporates retrieved medical content to guide its responses.

\section{Experimental Setup}

We evaluate our proposed \model\ model through both expert human evaluation and an LLM-as-Judge protocol. Our goal is to assess the clinical quality of responses generated by various model configurations, focusing on three key axes: factual accuracy, clinical relevance, and comprehensiveness, in the context of bariatric surgery patient education.

We benchmark four categories of systems: (i) \textbf{Offline RAG}, which includes a standard retrieval-augmented generation baseline and a domain-tuned MedGraphRAG variant using graph-based retrieval; (ii) \textbf{Zero-shot LLMs}, where large language models (Llama3-8B, Phi-3, and Mistral Instruct) generate answers without additional context; (iii) \textbf{Context-Prompted LLMs}, where retrieved context is appended at inference time without parameter updates; and (iv) \textbf{\model\ (Proposed)}, which applies confidence-aware parametric updates using retrieved evidence. All systems are tested under identical conditions and use a shared retrieval pipeline where applicable.

For expert evaluation, we consult one board-certified bariatric surgeon, who reviewed model outputs for $105$ instances. Each response is scored independently across three dimensions: \textbf{Factuality} (accuracy and correctness), \textbf{Clinical Relevance} (appropriateness in a clinical context), and \textbf{Comprehensiveness} (completeness and informativeness for patients). Ratings are based on a 5-point Likert scale, with final scores reported as averages across questions.

To complement the expert review and enable scalable comparison, we also evaluate all models using an \textbf{LLM-as-Judge} setup, where ChatGPT-4o\footnote{https://openai.com/index/hello-gpt-4o/} is prompted with each question, the corresponding model-generated answer, and a rubric defining the evaluation criteria. The model then rates each answer on the same 1–5 scale. To assess the reliability of this proxy, we compute the rank correlation between expert and ChatGPT-4o scores and observe a strong alignment ($\rho = 0.94$), confirming the viability of using LLMs for early-stage model quality assessment.

\begin{table}[!htp]
\centering
\small
\begin{tabular}{llrrrr}
\toprule
\multicolumn{2}{c}{\textbf{System}} & \multicolumn{4}{c}{\textbf{Metrics}} \\
\cmidrule{1-6}
\textbf{Type} & \textbf{Model} & \textbf{Factuality} & \textbf{Clinical Rel.} & \textbf{Compre.} & \textbf{Avg} \\
\midrule
Offline & RAG$^2$ & 3.62 & 3.45 & 3.53 & 3.53 \\
 & MedGraphRAG & 3.85 & 3.92 & 4.38 & 4.05 \\
 \hdashline
Zero-shot & Llama3-8B & 3.41 & 3.25 & 3.46 & 3.37 \\
 & Phi-3 & 2.37 & 2.15 & 2.25 & 2.26 \\
 & Mistral instruct & 2.23 & 2.18 & 2.14 & 2.18 \\
 \hdashline
Context prompted & Llama3-8B & 3.82 & 3.91 & 4.34 & 4.02 \\
 & Phi-3 & 2.64 & 2.75 & 2.42 & 2.60 \\
 & Mistral instruct & 2.69 & 2.37 & 3.35 & 2.80 \\
 \hdashline
\textbf{\model\ \textit{with}} & Llama3-8B & \textbf{4.18}  & \textbf{4.58}  & \textbf{4.76}  & \textbf{4.51}  \\
 & Phi-3 & \textbf{2.87}  & \textbf{3.17}  & \textbf{2.60}  & \textbf{2.88}  \\
 & Mistral instruct & \textbf{2.95}  & \textbf{2.71}  & \textbf{3.63}  & \textbf{3.09}  \\
\bottomrule
\end{tabular}
\caption{
Evaluation of various models across different configurations by expert surgeons. Each system is evaluated on Factuality, Clinical Relevance (Clinical Rel.), and Comprehensiveness (Compre.), with scores from 1 (poor) to 5 (excellent). The final three rows under each model group show the performance of our Online \model\ setting. The 'Avg' column reports the average of the three evaluation metrics.}
\label{tab:expert_judge_evaluation}
\end{table}

\section{Results}

We evaluate our proposed model, \model, using both expert evaluation review by board-certified bariatric surgeon and LLM-as-Judge using ChatGPT-4o. We compare \model\ against a suite of baselines, including standard retrieval-augmented models, zero-shot LLMs, and context-prompted variants.

\subsection{Expert Evaluation}

To assess the clinical quality of generated responses, we conducted a blinded evaluation with board-certified bariatric surgeon, who rated model outputs across three dimensions: \textbf{Factuality}, \textbf{Clinical Relevance}, and \textbf{Comprehensiveness}, using a 1--5 Likert scale (higher is better). Table~\ref{tab:expert_judge_evaluation} presents the average scores for each system under multiple configurations.

Among all baselines, \textbf{MedGraphRAG}, an offline domain-specific RAG model, achieved the highest average score (4.05), outperforming both standard offline RAG baselines (RAG$^2$: 3.53) and all zero-shot models (Llama3-8B: 3.37; Phi-3: 2.26; Mistral: 2.18). Context-prompted models (i.e., inputting relevant question context during inference) moderately improved scores, especially for Llama3-8B (Avg: 4.02), but still fell short of delivering optimal factual and clinical consistency.

Our proposed \model\ framework delivered the best overall performance across all metrics. When paired with Llama3-8B, \model\ achieved the highest average score (4.51), with near-expert level performance on \textbf{Comprehensiveness} (4.76) and \textbf{Clinical Relevance} (4.58). Notably, \model\ also improved the performance of smaller models like Phi-3 and Mistral, elevating their average scores by +0.6--0.7 points compared to their context-prompted or zero-shot baselines. These gains highlight the effectiveness of our confidence-aware updating mechanism, which not only retrieves up-to-date clinical evidence but also integrates it into the model’s internal parameters, enabling more robust, domain-adapted reasoning.

These results demonstrate that \model\ significantly enhances the clinical utility of LLMs across model sizes, especially when compared to conventional static RAG setups or prompting-only strategies.

\begin{table}[!htp]
\small
\centering
\begin{tabular}{llrrrr}
\toprule
\multicolumn{2}{c}{\textbf{System}} & \multicolumn{4}{c}{\textbf{Metrics}} \\
\cmidrule{1-6}
\textbf{Type} & \textbf{Model} & \textbf{Factuality} & \textbf{Clinical Rel.} & \textbf{Compre.} & \textbf{Avg} \\
\midrule
Offline & RAG$^2$ & 3.49 & 3.28 & 3.36 & 3.38 \\
 & MedGraphRAG & 3.67 & 3.76 & 4.45 & 3.96 \\
 \hdashline
Zero-shot & Llama3-8B & 3.28 & 3.18 & 3.34 & 3.27 \\
 & Phi-3 & 2.24 & 2.07 & 2.12 & 2.14 \\
 & Mistral instruct & 2.17 & 2.11 & 2.01 & 2.10 \\
 \hdashline
Context prompted & Llama3-8B & 3.67 & 3.76 & 4.45 & 3.96 \\
 & Phi-3 & 2.49 & 2.61 & 2.38 & 2.49 \\
 & Mistral instruct & 2.57 & 2.24 & 3.21 & 2.67 \\
 \hdashline
\textbf{\model\ \textit{with}} & Llama3-8B & \textbf{4.03}  & \textbf{4.43}  & \textbf{4.87}  & \textbf{4.44}  \\
 & Phi-3 & \textbf{2.73}  & \textbf{3.03}  & \textbf{2.54}  & \textbf{2.77}  \\
 & Mistral instruct & \textbf{2.83}  & \textbf{2.58}  & \textbf{3.48}  & \textbf{2.96}  \\
\bottomrule
\end{tabular}
\caption{
Evaluation of various models across different configurations using LLM-as-Judge metrics. Each system is evaluated on Factuality, Clinical Relevance (Clinical Rel.), and Comprehensiveness (Compre.), with scores from 1 (poor) to 5 (excellent). The final three rows under each model group show the performance of our Online \model\ setting. The 'Avg' column reports the average of the three evaluation metrics.}
\vspace{-6mm}
\label{tab:llm_judge_evaluation}
\end{table}

\subsection{LLM-as-Judge Evaluation}

To complement expert evaluation, we further assess all models using an LLM-as-Judge framework, where we use ChatGPT-4o model scores responses along three axes: \textbf{Factuality}, \textbf{Clinical Relevance}, and \textbf{Comprehensiveness}, using a 5-point Likert scale. Table~\ref{tab:llm_judge_evaluation} summarizes the performance of baseline systems and our proposed \model\ across these dimensions.

Among the baselines, MedGraphRAG and context-prompted Llama3-8B show relatively strong performance, achieving average scores of $3.96$. However, our proposed \model\ approach consistently outperforms all baselines across all metrics and models. For instance, \model\ with Llama3-8B achieves the highest overall score of $4.44$, reflecting substantial improvements in factual correctness ($+0.36$), clinical relevance ($+0.67$), and comprehensiveness ($+0.42$) over the best-performing baseline.

Notably, even with smaller models like Phi-3 and Mistral instruct, \model\ enhances output quality, particularly in relevance and completeness. These findings demonstrate that our system not only boosts performance for high-capacity LLMs but also meaningfully improves the reliability of lightweight models, making it practical for resource-constrained settings.

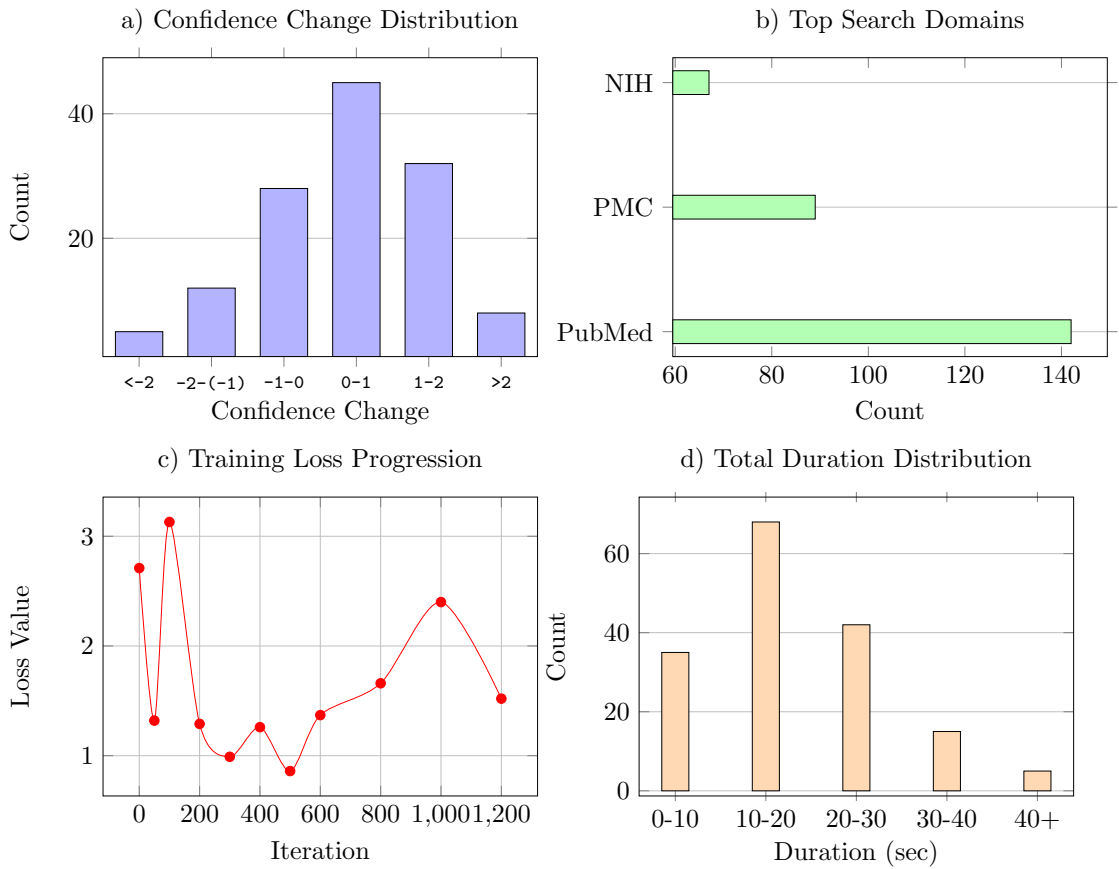
\begin{figure}[t]
\centering
\begin{tikzpicture}
\scalebox{0.9}{
\begin{axis}[
    title={ a) Confidence Change Distribution},
    width=8cm,
    height=6cm,
    ybar,
    xlabel={Confidence Change},
    ylabel={Count},
    xtick={-3,-2,-1,0,1,2},
   xticklabels={
  {\scriptsize\texttt{<}\newline\texttt{-2}},
  {\scriptsize\texttt{-2}\newline\texttt{-(-1)}},
  {\scriptsize\texttt{-1}\newline\texttt{-0}},
  {\scriptsize\texttt{0}\newline\texttt{-1}},
  {\scriptsize\texttt{1}\newline\texttt{-2}},
  {\scriptsize\texttt{>}\newline\texttt{2}}
},
    ymajorgrids=true,
    bar width=7mm,
    name=plot1
]
\addplot[fill=blue!30] coordinates {
    (-3,5) (-2,12) (-1,28) (0,45) (1,32) (2,8)
};
\end{axis}}

\scalebox{0.9}{
\begin{axis}[
    title={ b) Top Search Domains},
    width=8cm,
    height=6cm,
    xbar,
    xlabel={Count},
    ytick=data,
    yticklabels={PubMed, PMC, NIH},
    ymajorgrids=true,
    at={(plot1.east)},
    anchor=west,
    xshift=2cm
]
\addplot[fill=green!30] coordinates {
    (142,0)
    (89,1)
    (67,2)
};
\end{axis}}
\end{tikzpicture}

\vspace{-0.6cm} 

\begin{tikzpicture}
\scalebox{0.9}{
\begin{axis}[
    title={ c) Training Loss Progression},
    width=8cm,
    height=6cm,
    xlabel={Iteration},
    ylabel={Loss Value},
    grid=major,
    name=plot3
]
\addplot[color=red,mark=*,smooth] coordinates {
    (0,2.71) (50,1.32) (100,3.13) (200,1.29) (300,0.99) 
    (400,1.26) (500,0.86) (600,1.37) (800,1.66) 
    (1000,2.40) (1200,1.52)
};
\end{axis}}

\scalebox{0.9}{
\begin{axis}[
    title={ d) Total Duration Distribution},
    width=8cm,
    height=6cm,
    ybar,
    xlabel={Duration (sec)},
    ylabel={Count},
    ymajorgrids=true,
    bar width=4mm,
    at={(plot3.east)},
    anchor=west,
    xshift=1.5cm,
    xtick={5,15,25,35,45},  
    xticklabels={0-10,10-20,20-30,30-40,40+},  
    yticklabel style={align=left},  
    xtick align=center  
]
\addplot[fill=orange!30] coordinates {
    (5,35) (15,68) (25,42) (35,15) (45,5)
};
\end{axis}}

\end{tikzpicture}

\caption{Exploratory Analysis of Model Editing Dynamics.
(a) Distribution of changes in confidence scores post-edit, showing that most changes are modest and positive. 
(b) Frequency of search queries across external biomedical domains, with PubMed dominating.
(c) Training loss progression across iterations, illustrating convergence patterns and volatility.
(d) Distribution of total duration taken for each edit operation, highlighting that most edits are executed within 10-20 seconds.
}
\vspace{-7mm}
\label{fig:stats}
\end{figure}

\subsection{Expert vs. LLM-as-Judge: Score Alignment}
\label{sec:alignment}
To assess the alignment between human and expert evaluation, we compare expert ratings (Table~\ref{tab:expert_judge_evaluation}) with those produced by the LLM-as-Judge framework (Table~\ref{tab:llm_judge_evaluation}) across all models and configurations.

Overall, we observe a high degree of consistency in relative rankings across systems. For instance, both experts and the LLM-as-Judge identify MedGraphRAG and context-prompted Llama3-8B as the strongest baselines, while zero-shot models like Phi-3 and Mistral perform the worst across all axes. Furthermore, our proposed \model\ yields the highest scores in both evaluation schemes, affirming its robustness across human and model-based judgments.

Metric-wise, the strongest agreement is seen in the \textbf{Comprehensiveness} and \textbf{Clinical Relevance} dimensions, where score trends closely track each other across settings. Some minor variation arises in the \textbf{Factuality} scores, particularly for models like Phi-3 and Mistral, where the LLM-as-Judge is slightly more conservative than human reviewers. This discrepancy likely stems from the LLM's heightened sensitivity to surface-level inaccuracies, compared to domain experts who may weigh overall clinical soundness more heavily.

Importantly, the average correlation between expert and LLM-as-Judge scores across all models is $\rho = 0.94$ (Spearman), underscoring the reliability of using LLMs as surrogate evaluators in low-resource or iterative development settings. These results suggest that LLM-as-Judge provides a scalable and reasonably aligned proxy for expert review—particularly useful for rapid benchmarking and ablation testing during system development.

\section{\model\ Analysis}

Figure~\ref{fig:stats} presents a comprehensive analysis across multiple dimensions to evaluate the behavior, responsiveness, and efficiency of \model\ during real-time knowledge integration in the context of bariatric care.

\textbf{(a) Confidence Change Distribution.} The histogram in Fig. \ref{fig:stats}.a illustrates the distribution of confidence score changes triggered by the adaptive retrieval mechanism. Most examples exhibit moderate confidence gains (bins \texttt{0-1} and \texttt{1-2}), with the highest concentration in the \texttt{0-1} bin. This confirms that the system’s dynamic thresholding yields frequent yet stable updates, allowing the model to autonomously improve responses without overcorrecting. Rare occurrences of extreme confidence shifts (\texttt{<-2} or \texttt{>2}) indicate that the system maintains a conservative stance, prioritizing stability in medical contexts.

\textbf{(b) Top Search Domains.} To assess external evidence sources, we analyzed the frequency of domain-level API queries. \texttt{PubMed}, \texttt{PMC}, and \texttt{NIH} emerged as the top knowledge sources (cf. Fig \ref{fig:stats}.b), underscoring \model's strong preference for authoritative biomedical repositories. This supports the design goal of maintaining clinical fidelity and alignment with evidence-based guidelines during patient-facing interactions.

\textbf{(c) Training Loss Progression.} The loss trajectory (cf. Fig \ref{fig:stats}.c) across 1,302 iterations reveals several key phases in the model’s learning dynamics. An initial sharp drop from $2.71$ to $1.32$ by iteration $50$ is followed by a spike to $3.13$ at iteration $100$, likely due to early exploratory updates. Subsequent iterations show improved stability and convergence, with the lowest loss ($0.86$) reached around iteration $500$. Notable local minima at iterations $300$ and $600$ indicate consistent refinement, while the uptick at iteration $1000$ may reflect a transient deviation before re-stabilization. Overall, the pattern confirms that \model's adaptive updating mechanism supports gradual convergence while accommodating knowledge volatility.

\textbf{(d) Total Duration Distribution.} The majority of update operations complete within \texttt{10-20 seconds}, with fewer cases extending beyond \texttt{30 seconds}. This distribution validates that \model’s self-updating pipeline is both computationally lightweight and suitable for real-time deployment in longitudinal MBS care settings ensuring timely and trustworthy information delivery across all stages of the surgical journey.

Further qualitative comparisons across diverse questions and model outputs are presented in Appendix~\ref{appendix:analysis}.



\section{Conclusion}
We introduced \model, an adaptive retrieval-augmented generation (RAG) system tailored for bariatric and metabolic surgery (MBS) support. By autonomously incorporating real-time medical evidence when confidence dips below dynamic thresholds, \model\ ensures that responses remain timely, accurate, and clinically reliable. To facilitate robust benchmarking, we also introduced \data, the first large-scale, expert-validated dataset of postoperative bariatric care questions. Through comprehensive evaluation using both LLM-based metrics and expert surgeon assessments, \model\ consistently outperformed existing state-of-the-art models in clinical accuracy and relevance.


\bibliography{main}

\begin{thebibliography}{52}
\providecommand{\natexlab}[1]{#1}
\providecommand{\url}[1]{\texttt{#1}}
\expandafter\ifx\csname urlstyle\endcsname\relax
  \providecommand{\doi}[1]{doi: #1}\else
  \providecommand{\doi}{doi: \begingroup \urlstyle{rm}\Url}\fi

\bibitem[Abdin et~al.(2024)Abdin, Aneja, Awadalla, and et~al.]{abdin2024phi3technicalreporthighly}
Marah Abdin, Jyoti Aneja, Hany Awadalla, and et~al.
\newblock Phi-3 technical report: A highly capable language model locally on your phone, 2024.
\newblock URL \url{https://arxiv.org/abs/2404.14219}.

\bibitem[Alber et~al.(2025)Alber, Yang, Alyakin, Yang, Rai, Valliani, Zhang, Rosenbaum, Amend-Thomas, Kurland, Kremer, Eremiev, Negash, Wiggan, Nakatsuka, Sangwon, Neifert, Khan, Save, Palla, Grin, Hedman, Nasir-Moin, Liu, Jiang, Mankowski, Segev, Aphinyanaphongs, Riina, Golfinos, Orringer, Kondziolka, and Oermann]{Alber2025-qb}
Daniel~Alexander Alber, Zihao Yang, Anton Alyakin, Eunice Yang, Sumedha Rai, Aly~A Valliani, Jeff Zhang, Gabriel~R Rosenbaum, Ashley~K Amend-Thomas, David~B Kurland, Caroline~M Kremer, Alexander Eremiev, Bruck Negash, Daniel~D Wiggan, Michelle~A Nakatsuka, Karl~L Sangwon, Sean~N Neifert, Hammad~A Khan, Akshay~Vinod Save, Adhith Palla, Eric~A Grin, Monika Hedman, Mustafa Nasir-Moin, Xujin~Chris Liu, Lavender~Yao Jiang, Michal~A Mankowski, Dorry~L Segev, Yindalon Aphinyanaphongs, Howard~A Riina, John~G Golfinos, Daniel~A Orringer, Douglas Kondziolka, and Eric~Karl Oermann.
\newblock Medical large language models are vulnerable to data-poisoning attacks.
\newblock \emph{Nat. Med.}, 31\penalty0 (2):\penalty0 618--626, February 2025.

\bibitem[Atri et~al.(2021)Atri, Pramanick, Goyal, and Chakraborty]{ATRI2021107152}
Yash~Kumar Atri, Shraman Pramanick, Vikram Goyal, and Tanmoy Chakraborty.
\newblock See, hear, read: Leveraging multimodality with guided attention for abstractive text summarization.
\newblock \emph{Knowledge-Based Systems}, 227:\penalty0 107152, 2021.
\newblock ISSN 0950-7051.
\newblock \doi{https://doi.org/10.1016/j.knosys.2021.107152}.
\newblock URL \url{https://www.sciencedirect.com/science/article/pii/S0950705121004159}.

\bibitem[Atri et~al.(2023{\natexlab{a}})Atri, Goyal, and Chakraborty]{10.1145/3580305.3599830}
Yash~Kumar Atri, Vikram Goyal, and Tanmoy Chakraborty.
\newblock Fusing multimodal signals on hyper-complex space for extreme abstractive text summarization (tl;dr) of scientific contents.
\newblock In \emph{Proceedings of the 29th ACM SIGKDD Conference on Knowledge Discovery and Data Mining}, KDD '23, page 3724–3736, New York, NY, USA, 2023{\natexlab{a}}. Association for Computing Machinery.
\newblock ISBN 9798400701030.
\newblock \doi{10.1145/3580305.3599830}.
\newblock URL \url{https://doi.org/10.1145/3580305.3599830}.

\bibitem[Atri et~al.(2023{\natexlab{b}})Atri, Goyal, and Chakraborty]{10261260}
Yash~Kumar Atri, Vikram Goyal, and Tanmoy Chakraborty.
\newblock Multi-document summarization using selective attention span and reinforcement learning.
\newblock \emph{IEEE/ACM Transactions on Audio, Speech, and Language Processing}, 31:\penalty0 3457--3467, 2023{\natexlab{b}}.
\newblock \doi{10.1109/TASLP.2023.3316459}.

\bibitem[Atri et~al.(2023{\natexlab{c}})Atri, Iyer, Chakraborty, and Goyal]{atri-etal-2023-promoting}
Yash~Kumar Atri, Arun Iyer, Tanmoy Chakraborty, and Vikram Goyal.
\newblock Promoting topic coherence and inter-document consorts in multi-document summarization via simplicial complex and sheaf graph.
\newblock In Houda Bouamor, Juan Pino, and Kalika Bali, editors, \emph{Proceedings of the 2023 Conference on Empirical Methods in Natural Language Processing}, pages 2154--2166, Singapore, December 2023{\natexlab{c}}. Association for Computational Linguistics.
\newblock \doi{10.18653/v1/2023.emnlp-main.133}.
\newblock URL \url{https://aclanthology.org/2023.emnlp-main.133/}.

\bibitem[Barres et~al.(2013)Barres, Kirchner, Rasmussen, Yan, Kantor, Krook, Naslund, Zierath, and Ling]{Barres2013}
Romain Barres, Henri Kirchner, Morten Rasmussen, Jing Yan, Daniel Kantor, Anna Krook, Erik Naslund, Juleen~R. Zierath, and Charlotte Ling.
\newblock Weight loss after gastric bypass surgery in human obesity remodels promoter methylation.
\newblock \emph{Cell Reports}, 3\penalty0 (4):\penalty0 1020--1027, 2013.
\newblock \doi{10.1016/j.celrep.2013.03.019}.

\bibitem[Bartholomay et~al.(2024)Bartholomay, Stewart, Sarwer, Wadden, and Fabricatore]{Bartholomay2024}
Emily~M. Bartholomay, Patrick~W. Stewart, David~B. Sarwer, Thomas~A. Wadden, and Anthony~N. Fabricatore.
\newblock Sociodemographic factors related to bariatric follow-up appointment attendance and weight outcomes.
\newblock \emph{Surgery for Obesity and Related Diseases}, 20:\penalty0 1388--1395, 2024.
\newblock \doi{10.1016/j.soard.2024.02.010}.

\bibitem[B{\'e}lisle-Pipon(2024)]{Belisle-Pipon2024-pj}
Jean-Christophe B{\'e}lisle-Pipon.
\newblock Why we need to be careful with {LLMs} in medicine.
\newblock \emph{Front. Med. (Lausanne)}, 11:\penalty0 1495582, December 2024.

\bibitem[Bjerkan et~al.(2022)Bjerkan, Viste, Aasheim, Mjåland, Mala, Kløw, Røislien, and Bøhn]{Bjerkan2022}
Kristin~K. Bjerkan, Audun Viste, Else~M. Aasheim, Oda Mjåland, Torstein Mala, Nina~E. Kløw, Jo~Røislien, and Siv~K. Bøhn.
\newblock The long-term impact of postoperative educational programs on weight loss after roux-en-y gastric bypass.
\newblock \emph{Obesity Surgery}, 32:\penalty0 3005--3012, 2022.
\newblock \doi{10.1007/s11695-022-05913-0}.

\bibitem[Cai et~al.(2024)Cai, Tan, Song, Sun, Jiang, Xu, Zhang, and Gu]{Cai_2024}
Tianchi Cai, Zhiwen Tan, Xierui Song, Tao Sun, Jiyan Jiang, Yunqi Xu, Yinger Zhang, and Jinjie Gu.
\newblock Forag: Factuality-optimized retrieval augmented generation for web-enhanced long-form question answering.
\newblock In \emph{Proceedings of the 30th ACM SIGKDD Conference on Knowledge Discovery and Data Mining}, KDD ’24, page 199–210. ACM, August 2024.
\newblock \doi{10.1145/3637528.3672065}.
\newblock URL \url{http://dx.doi.org/10.1145/3637528.3672065}.

\bibitem[Cheng et~al.(2024)Cheng, Marone, Weller, Lawrie, Khashabi, and Durme]{cheng2024dateddatatracingknowledge}
Jeffrey Cheng, Marc Marone, Orion Weller, Dawn Lawrie, Daniel Khashabi, and Benjamin~Van Durme.
\newblock Dated data: Tracing knowledge cutoffs in large language models, 2024.
\newblock URL \url{https://arxiv.org/abs/2403.12958}.

\bibitem[Clapp et~al.(2024)Clapp, Khaitan, English, Gagner, Inabnet, Morton, Pories, Schauer, Wolfe, and Wolfe]{Clapp2024}
Bryan Clapp, Lillian Khaitan, William~J. English, Michel Gagner, William~B. Inabnet, J.~Michael Morton, Walter~J. Pories, Philip~R. Schauer, Brian~M. Wolfe, and Mary~M. Wolfe.
\newblock American society for metabolic and bariatric surgery 2022 estimate of metabolic and bariatric procedures performed in the united states.
\newblock \emph{Surgery for Obesity and Related Diseases}, 20:\penalty0 425--431, 2024.
\newblock \doi{10.1016/j.soard.2023.11.005}.

\bibitem[David et~al.(2020)David, Sijercic, and Cassin]{David2020}
Lauren~A. David, Ivana Sijercic, and Stephanie~E. Cassin.
\newblock Preoperative and post-operative psychosocial interventions for bariatric surgery patients: A systematic review.
\newblock \emph{Obesity Reviews}, 21:\penalty0 e12926, 2020.
\newblock \doi{10.1111/obr.12926}.

\bibitem[Dey et~al.(2020)Dey, Chowdhury, Atri, and Chakraborty]{dey-etal-2020-corpora}
Alvin Dey, Tanya Chowdhury, Yash~Kumar Atri, and Tanmoy Chakraborty.
\newblock Corpora evaluation and system bias detection in multi-document summarization.
\newblock In Trevor Cohn, Yulan He, and Yang Liu, editors, \emph{Findings of the Association for Computational Linguistics: EMNLP 2020}, pages 2830--2840, Online, November 2020. Association for Computational Linguistics.
\newblock \doi{10.18653/v1/2020.findings-emnlp.254}.
\newblock URL \url{https://aclanthology.org/2020.findings-emnlp.254/}.

\bibitem[Gao et~al.(2024)Gao, Xiong, Gao, Jia, Pan, Bi, Dai, Sun, Wang, and Wang]{gao2024retrievalaugmentedgenerationlargelanguage}
Yunfan Gao, Yun Xiong, Xinyu Gao, Kangxiang Jia, Jinliu Pan, Yuxi Bi, Yi~Dai, Jiawei Sun, Meng Wang, and Haofen Wang.
\newblock Retrieval-augmented generation for large language models: A survey, 2024.
\newblock URL \url{https://arxiv.org/abs/2312.10997}.

\bibitem[Grattafiori et~al.(2024)Grattafiori, Dubey, Jauhri, and et~al.]{grattafiori2024llama3herdmodels}
Aaron Grattafiori, Abhimanyu Dubey, Abhinav Jauhri, and et~al.
\newblock The llama 3 herd of models, 2024.
\newblock URL \url{https://arxiv.org/abs/2407.21783}.

\bibitem[Groller et~al.(2017)Groller, Carey, and Evans]{Groller2017}
Kristen~D. Groller, Kasey~M. Carey, and John~P. Evans.
\newblock Systematic review of patient education practices in weight loss surgery.
\newblock \emph{Surgery for Obesity and Related Diseases}, 13:\penalty0 1072--1085, 2017.
\newblock \doi{10.1016/j.soard.2017.02.025}.

\bibitem[Heydari et~al.(2025)Heydari, Hemmat, Naman, and Fatemi]{heydari2025contextawarenessgateretrieval}
Mohammad~Hassan Heydari, Arshia Hemmat, Erfan Naman, and Afsaneh Fatemi.
\newblock Context awareness gate for retrieval augmented generation, 2025.
\newblock URL \url{https://arxiv.org/abs/2411.16133}.

\bibitem[Javanparast et~al.(2021)Javanparast, Roeger, Kwok, and Reed]{Javanparast2021-te}
Sara Javanparast, Leigh Roeger, Yuen Kwok, and Richard~L Reed.
\newblock The experience of australian general practice patients at high risk of poor health outcomes with telehealth during the {COVID-19} pandemic: a qualitative study.
\newblock \emph{BMC Fam. Pract.}, 22\penalty0 (1):\penalty0 69, April 2021.

\bibitem[Jiang et~al.(2023)Jiang, Sablayrolles, Mensch, Bamford, Chaplot, de~las Casas, Bressand, Lengyel, Lample, Saulnier, Lavaud, Lachaux, Stock, Scao, Lavril, Wang, Lacroix, and Sayed]{jiang2023mistral7b}
Albert~Q. Jiang, Alexandre Sablayrolles, Arthur Mensch, Chris Bamford, Devendra~Singh Chaplot, Diego de~las Casas, Florian Bressand, Gianna Lengyel, Guillaume Lample, Lucile Saulnier, Lélio~Renard Lavaud, Marie-Anne Lachaux, Pierre Stock, Teven~Le Scao, Thibaut Lavril, Thomas Wang, Timothée Lacroix, and William~El Sayed.
\newblock Mistral 7b, 2023.
\newblock URL \url{https://arxiv.org/abs/2310.06825}.

\bibitem[Jiang et~al.(2025)Jiang, Sun, Liang, and Zhang]{jiang2025retrievesummarizeplanadvancing}
Zhouyu Jiang, Mengshu Sun, Lei Liang, and Zhiqiang Zhang.
\newblock Retrieve, summarize, plan: Advancing multi-hop question answering with an iterative approach, 2025.
\newblock URL \url{https://arxiv.org/abs/2407.13101}.

\bibitem[Jin et~al.(2019)Jin, Dhingra, Liu, Cohen, and Lu]{jin-etal-2019-pubmedqa}
Qiao Jin, Bhuwan Dhingra, Zhengping Liu, William Cohen, and Xinghua Lu.
\newblock {P}ub{M}ed{QA}: A dataset for biomedical research question answering.
\newblock In Kentaro Inui, Jing Jiang, Vincent Ng, and Xiaojun Wan, editors, \emph{Proceedings of the 2019 Conference on Empirical Methods in Natural Language Processing and the 9th International Joint Conference on Natural Language Processing (EMNLP-IJCNLP)}, pages 2567--2577, Hong Kong, China, November 2019. Association for Computational Linguistics.
\newblock \doi{10.18653/v1/D19-1259}.
\newblock URL \url{https://aclanthology.org/D19-1259/}.

\bibitem[Khlaut et~al.(2024)Khlaut, Dancette, Ferreres, Alaedine, Herent, and Manceron]{khlaut-etal-2024-efficient}
Julien Khlaut, Corentin Dancette, Elodie Ferreres, Benani Alaedine, Herent Herent, and Pierre Manceron.
\newblock Efficient medical question answering with knowledge-augmented question generation.
\newblock In Tristan Naumann, Asma Ben~Abacha, Steven Bethard, Kirk Roberts, and Danielle Bitterman, editors, \emph{Proceedings of the 6th Clinical Natural Language Processing Workshop}, pages 10--20, Mexico City, Mexico, June 2024. Association for Computational Linguistics.
\newblock \doi{10.18653/v1/2024.clinicalnlp-1.2}.
\newblock URL \url{https://aclanthology.org/2024.clinicalnlp-1.2/}.

\bibitem[Kim et~al.(2023)Kim, Davenport, Gu, Ziegler, Ferguson, and Farrell]{Kim2023}
Dong~H. Kim, Melissa~S. Davenport, Jia Gu, Kristen Ziegler, Patricia~L. Ferguson, and Timothy~M. Farrell.
\newblock Incidence, burden, and predictors of 11-month readmission in patients undergoing bariatric surgery.
\newblock \emph{Obesity Surgery}, 33:\penalty0 94--104, 2023.
\newblock \doi{10.1007/s11695-022-06410-2}.

\bibitem[Kim et~al.(2024)Kim, Wu, Abdulle, and Wu]{kim-etal-2024-medexqa}
Yunsoo Kim, Jinge Wu, Yusuf Abdulle, and Honghan Wu.
\newblock {M}ed{E}x{QA}: Medical question answering benchmark with multiple explanations.
\newblock In Dina Demner-Fushman, Sophia Ananiadou, Makoto Miwa, Kirk Roberts, and Junichi Tsujii, editors, \emph{Proceedings of the 23rd Workshop on Biomedical Natural Language Processing}, pages 167--181, Bangkok, Thailand, August 2024. Association for Computational Linguistics.
\newblock \doi{10.18653/v1/2024.bionlp-1.14}.
\newblock URL \url{https://aclanthology.org/2024.bionlp-1.14/}.

\bibitem[Kojima et~al.(2023)Kojima, Gu, Reid, Matsuo, and Iwasawa]{kojima2023largelanguagemodelszeroshot}
Takeshi Kojima, Shixiang~Shane Gu, Machel Reid, Yutaka Matsuo, and Yusuke Iwasawa.
\newblock Large language models are zero-shot reasoners, 2023.
\newblock URL \url{https://arxiv.org/abs/2205.11916}.

\bibitem[Lewis et~al.(2020)Lewis, Perez, Piktus, Petroni, Karpukhin, Goyal, K\"{u}ttler, Lewis, Yih, Rockt\"{a}schel, Riedel, and Kiela]{10.5555/3495724.3496517}
Patrick Lewis, Ethan Perez, Aleksandra Piktus, Fabio Petroni, Vladimir Karpukhin, Naman Goyal, Heinrich K\"{u}ttler, Mike Lewis, Wen-tau Yih, Tim Rockt\"{a}schel, Sebastian Riedel, and Douwe Kiela.
\newblock Retrieval-augmented generation for knowledge-intensive nlp tasks.
\newblock In \emph{Proceedings of the 34th International Conference on Neural Information Processing Systems}, NIPS '20, Red Hook, NY, USA, 2020. Curran Associates Inc.
\newblock ISBN 9781713829546.

\bibitem[Lewis et~al.(2021)Lewis, Perez, Piktus, Petroni, Karpukhin, Goyal, Küttler, Lewis, tau Yih, Rocktäschel, Riedel, and Kiela]{lewis2021retrievalaugmentedgenerationknowledgeintensivenlp}
Patrick Lewis, Ethan Perez, Aleksandra Piktus, Fabio Petroni, Vladimir Karpukhin, Naman Goyal, Heinrich Küttler, Mike Lewis, Wen tau Yih, Tim Rocktäschel, Sebastian Riedel, and Douwe Kiela.
\newblock Retrieval-augmented generation for knowledge-intensive nlp tasks, 2021.
\newblock URL \url{https://arxiv.org/abs/2005.11401}.

\bibitem[Li et~al.(2024)Li, Yuan, and Zhang]{li2024enhancingllmfactualaccuracy}
Jiarui Li, Ye~Yuan, and Zehua Zhang.
\newblock Enhancing llm factual accuracy with rag to counter hallucinations: A case study on domain-specific queries in private knowledge-bases, 2024.
\newblock URL \url{https://arxiv.org/abs/2403.10446}.

\bibitem[Loos and Yeo(2022)]{Loos2022}
Ruth J.~F. Loos and G.~S.~H. Yeo.
\newblock The genetics of obesity: from discovery to biology.
\newblock \emph{Nature Reviews Genetics}, 23:\penalty0 120--133, 2022.
\newblock \doi{10.1038/s41576-021-00414-z}.

\bibitem[Lu et~al.(2024)Lu, Zhao, and Wang]{lu-etal-2024-clinicalrag}
Yuxing Lu, Xukai Zhao, and Jinzhuo Wang.
\newblock {C}linical{RAG}: Enhancing clinical decision support through heterogeneous knowledge retrieval.
\newblock In Sha Li, Manling Li, Michael~JQ Zhang, Eunsol Choi, Mor Geva, Peter Hase, and Heng Ji, editors, \emph{Proceedings of the 1st Workshop on Towards Knowledgeable Language Models (KnowLLM 2024)}, pages 64--68, Bangkok, Thailand, August 2024. Association for Computational Linguistics.
\newblock \doi{10.18653/v1/2024.knowllm-1.6}.
\newblock URL \url{https://aclanthology.org/2024.knowllm-1.6/}.

\bibitem[McLennan et~al.(2023)McLennan, Kiran, McLeod, Li, Guo, and Birch]{McLennan2023}
Sean McLennan, Ravi~P. Kiran, Robert~S. McLeod, Victor Li, Yue Guo, and David~W. Birch.
\newblock Characteristics and outcomes for patients undergoing revisional bariatric surgery due to persistent obesity: a retrospective cohort study of 10,589 patients.
\newblock \emph{Surgical Endoscopy}, 37:\penalty0 4613--4622, 2023.
\newblock \doi{10.1007/s00464-022-09715-9}.

\bibitem[Mechanick et~al.(2020)Mechanick, Apovian, Brethauer, Garvey, Joffe, Kim, Kushner, Pender, Singh, of~Clinical~Endocrinologists, of~Endocrinology, Society, for Metabolic, Surgery, Association, and of~Anesthesiologists]{Mechanick2020}
Jeffrey~I. Mechanick, Caroline Apovian, Stephen Brethauer, W.~Timothy Garvey, Aaron~M. Joffe, Jill Kim, Robert~F. Kushner, John~R. Pender, Kalpana Singh, American~Association of~Clinical~Endocrinologists, American~College of~Endocrinology, The~Obesity Society, American~Society for Metabolic, Bariatric Surgery, Obesity~Medicine Association, and American~Society of~Anesthesiologists.
\newblock Clinical practice guidelines for the perioperative nutrition, metabolic, and nonsurgical support of patients undergoing bariatric procedures 2019 update.
\newblock \emph{Obesity (Silver Spring)}, 28:\penalty0 O1--O58, 2020.
\newblock \doi{10.1002/oby.22792}.

\bibitem[Minaee et~al.(2025)Minaee, Mikolov, Nikzad, Chenaghlu, Socher, Amatriain, and Gao]{minaee2025largelanguagemodelssurvey}
Shervin Minaee, Tomas Mikolov, Narjes Nikzad, Meysam Chenaghlu, Richard Socher, Xavier Amatriain, and Jianfeng Gao.
\newblock Large language models: A survey, 2025.
\newblock URL \url{https://arxiv.org/abs/2402.06196}.

\bibitem[P et~al.(2024)P, G, V, Chopra, and Emran]{P2024-sc}
Jeyanthi P, Gulothungan G, Vinoth~Kumar V, Hitesh Chopra, and Talha~Bin Emran.
\newblock Enhancing postoperative care with telemedicine and remote monitoring for improved recovery and patient safety.
\newblock \emph{Int. J. Surg.}, 110\penalty0 (12):\penalty0 8205--8206, December 2024.

\bibitem[Patel and Thind(2020)]{Patel2020-xo}
Ben Patel and Arron Thind.
\newblock Usability of mobile health apps for postoperative care: Systematic review.
\newblock \emph{JMIR Perioper. Med.}, 3\penalty0 (2):\penalty0 e19099, July 2020.

\bibitem[Saeed(2024)]{saeed-2024-medifact}
Nadia Saeed.
\newblock {M}edi{F}act at {MEDIQA}-{M}3{G} 2024: Medical question answering in dermatology with multimodal learning.
\newblock In Tristan Naumann, Asma Ben~Abacha, Steven Bethard, Kirk Roberts, and Danielle Bitterman, editors, \emph{Proceedings of the 6th Clinical Natural Language Processing Workshop}, pages 339--345, Mexico City, Mexico, June 2024. Association for Computational Linguistics.
\newblock \doi{10.18653/v1/2024.clinicalnlp-1.31}.
\newblock URL \url{https://aclanthology.org/2024.clinicalnlp-1.31/}.

\bibitem[Schlottmann et~al.(2023)Schlottmann, Baz, Pirzada, and Masrur]{Schlottmann2023}
Francisco Schlottmann, Charlie Baz, Adeel Pirzada, and Muhammad~A. Masrur.
\newblock Postoperative follow-up compliance: The achilles’ heel of bariatric surgery.
\newblock \emph{Obesity Surgery}, 33:\penalty0 2945--2948, 2023.
\newblock \doi{10.1007/s11695-023-06500-2}.

\bibitem[Setarehdan et~al.(2023)Setarehdan, Ziaee, Niknam, Ghassemi, Mehrabi, Esfahani, and Shariat]{Setarehdan2023}
Saeed~A. Setarehdan, Ameneh Ziaee, Ramin Niknam, Sahar Ghassemi, Fatemeh Mehrabi, Mohsen~Hossein Esfahani, and Mohammad~H. Shariat.
\newblock Determinants of patient’s adherence to the predefined follow-up visits after bariatric surgery.
\newblock \emph{Obesity Surgery}, 33:\penalty0 577--584, 2023.
\newblock \doi{10.1007/s11695-022-06353-3}.

\bibitem[Singhal et~al.(2025)Singhal, Tu, Gottweis, Sayres, Wulczyn, Amin, Hou, Clark, Pfohl, Cole-Lewis, Neal, Rashid, Schaekermann, Wang, Dash, Chen, Shah, Lachgar, Mansfield, Prakash, Green, Dominowska, Ag{\"u}era Y~Arcas, Toma{\v s}ev, Liu, Wong, Semturs, Mahdavi, Barral, Webster, Corrado, Matias, Azizi, Karthikesalingam, and Natarajan]{Singhal2025-bf}
Karan Singhal, Tao Tu, Juraj Gottweis, Rory Sayres, Ellery Wulczyn, Mohamed Amin, Le~Hou, Kevin Clark, Stephen~R Pfohl, Heather Cole-Lewis, Darlene Neal, Qazi~Mamunur Rashid, Mike Schaekermann, Amy Wang, Dev Dash, Jonathan~H Chen, Nigam~H Shah, Sami Lachgar, Philip~Andrew Mansfield, Sushant Prakash, Bradley Green, Ewa Dominowska, Blaise Ag{\"u}era Y~Arcas, Nenad Toma{\v s}ev, Yun Liu, Renee Wong, Christopher Semturs, S~Sara Mahdavi, Joelle~K Barral, Dale~R Webster, Greg~S Corrado, Yossi Matias, Shekoofeh Azizi, Alan Karthikesalingam, and Vivek Natarajan.
\newblock Toward expert-level medical question answering with large language models.
\newblock \emph{Nat. Med.}, 31\penalty0 (3):\penalty0 943--950, March 2025.

\bibitem[Sohn et~al.(2024)Sohn, Park, Yoon, Park, Hwang, Sung, Kim, and Kang]{sohn2024rationaleguidedretrievalaugmentedgeneration}
Jiwoong Sohn, Yein Park, Chanwoong Yoon, Sihyeon Park, Hyeon Hwang, Mujeen Sung, Hyunjae Kim, and Jaewoo Kang.
\newblock Rationale-guided retrieval augmented generation for medical question answering, 2024.
\newblock URL \url{https://arxiv.org/abs/2411.00300}.

\bibitem[Sviridova et~al.(2024)Sviridova, Yeginbergen, Estarrona, Cabrio, Villata, and Agerri]{sviridova-etal-2024-casimedicos}
Ekaterina Sviridova, Anar Yeginbergen, Ainara Estarrona, Elena Cabrio, Serena Villata, and Rodrigo Agerri.
\newblock {C}asi{M}edicos-arg: A medical question answering dataset annotated with explanatory argumentative structures.
\newblock In Yaser Al-Onaizan, Mohit Bansal, and Yun-Nung Chen, editors, \emph{Proceedings of the 2024 Conference on Empirical Methods in Natural Language Processing}, pages 18463--18475, Miami, Florida, USA, November 2024. Association for Computational Linguistics.
\newblock \doi{10.18653/v1/2024.emnlp-main.1026}.
\newblock URL \url{https://aclanthology.org/2024.emnlp-main.1026/}.

\bibitem[Tamkin et~al.(2021)Tamkin, Brundage, Clark, and Ganguli]{tamkin2021understandingcapabilitieslimitationssocietal}
Alex Tamkin, Miles Brundage, Jack Clark, and Deep Ganguli.
\newblock Understanding the capabilities, limitations, and societal impact of large language models, 2021.
\newblock URL \url{https://arxiv.org/abs/2102.02503}.

\bibitem[Vladika and Matthes(2024)]{vladika-matthes-2024-improving}
Juraj Vladika and Florian Matthes.
\newblock Improving health question answering with reliable and time-aware evidence retrieval.
\newblock In Kevin Duh, Helena Gomez, and Steven Bethard, editors, \emph{Findings of the Association for Computational Linguistics: NAACL 2024}, pages 4752--4763, Mexico City, Mexico, June 2024. Association for Computational Linguistics.
\newblock \doi{10.18653/v1/2024.findings-naacl.295}.
\newblock URL \url{https://aclanthology.org/2024.findings-naacl.295/}.

\bibitem[Wang et~al.(2025)Wang, Zhang, He, Nguyen, Yu, Deng, Brandt, Bitterman, Pan, Cheng, Zou, and Liu]{wang2025safetychallengesaimedicine}
Xiaoye Wang, Nicole~Xi Zhang, Hongyu He, Trang Nguyen, Kun-Hsing Yu, Hao Deng, Cynthia Brandt, Danielle~S. Bitterman, Ling Pan, Ching-Yu Cheng, James Zou, and Dianbo Liu.
\newblock Safety challenges of ai in medicine in the era of large language models, 2025.
\newblock URL \url{https://arxiv.org/abs/2409.18968}.

\bibitem[Ward et~al.(2019)Ward, Bleich, Cradock, Barrett, Giles, Flax, and Long]{Ward2019}
Zachary~J. Ward, Sara~N. Bleich, Angie~L. Cradock, Jonathan~L. Barrett, Christine~M. Giles, Charles Flax, and Michael~W. Long.
\newblock Projected u.s. state-level prevalence of adult obesity and severe obesity.
\newblock \emph{The New England Journal of Medicine}, 381:\penalty0 2440--2450, 2019.
\newblock \doi{10.1056/NEJMsa1909301}.

\bibitem[Wu et~al.(2024{\natexlab{a}})Wu, Che, Zhang, Tao, Zhang, and Shao]{wu2024pandorasboxaladdinslamp}
Jinyang Wu, Feihu Che, Chuyuan Zhang, Jianhua Tao, Shuai Zhang, and Pengpeng Shao.
\newblock Pandora's box or aladdin's lamp: A comprehensive analysis revealing the role of rag noise in large language models, 2024{\natexlab{a}}.
\newblock URL \url{https://arxiv.org/abs/2408.13533}.

\bibitem[Wu et~al.(2024{\natexlab{b}})Wu, Zhu, Qi, Chen, Xu, Menolascina, and Grau]{wu2024medicalgraphragsafe}
Junde Wu, Jiayuan Zhu, Yunli Qi, Jingkun Chen, Min Xu, Filippo Menolascina, and Vicente Grau.
\newblock Medical graph rag: Towards safe medical large language model via graph retrieval-augmented generation, 2024{\natexlab{b}}.
\newblock URL \url{https://arxiv.org/abs/2408.04187}.

\bibitem[Wu et~al.(2024{\natexlab{c}})Wu, Graziano, Salner, Chen, Judge, Cong, and Xu]{Wu2024-nk}
Weizi Wu, Teresa Graziano, Andrew Salner, Ming-Hui Chen, Michelle~P Judge, Xiaomei Cong, and Wanli Xu.
\newblock Acceptability, effectiveness, and roles of mhealth applications in supporting cancer pain self-management: Integrative review.
\newblock \emph{JMIR MHealth UHealth}, 12:\penalty0 e53652, July 2024{\natexlab{c}}.

\bibitem[Zhang et~al.(2020)Zhang, Kishore, Wu, Weinberger, and Artzi]{zhang2020bertscoreevaluatingtextgeneration}
Tianyi Zhang, Varsha Kishore, Felix Wu, Kilian~Q. Weinberger, and Yoav Artzi.
\newblock Bertscore: Evaluating text generation with bert, 2020.
\newblock URL \url{https://arxiv.org/abs/1904.09675}.

\bibitem[Zhang and Zhang(2025)]{Zhang2025-wi}
Wan Zhang and Jing Zhang.
\newblock Hallucination mitigation for retrieval-augmented large language models: A review.
\newblock \emph{Mathematics}, 13\penalty0 (5):\penalty0 856, March 2025.

\end{thebibliography}

\newpage
\appendix

\section{Limitations and Future Directions}

While our proposed modular editing framework with LLaMA-3 demonstrates strong performance across factuality and faithfulness benchmarks, several limitations remain that open avenues for future research.

Scalability and Edit Interference. Although our method enables localized and compositional edits, the cumulative impact of many such edits—particularly in dense regions of the representation space—may lead to interference or capacity saturation. Future work could explore dynamic pruning or hierarchical edit graphs to manage long-term scalability.

Generalization Beyond Local Context. Our framework is effective at modifying behaviors related to specific factual updates. However, its ability to generalize edits to broader semantic or reasoning contexts is limited. Integrating structured world knowledge or training the modules with auxiliary objectives (e.g., counterfactual consistency) may further enhance generalization.

Evaluation Bottlenecks. While our automatic and human evaluations provide evidence of improved edit quality, assessing real-world applicability—especially in high-stakes domains—requires more robust metrics. Developing editing-specific benchmarks that reflect user intent, edit trustworthiness, and long-term retention would be a valuable direction.

Interpretability. Currently, our system relies on confidence thresholds and auxiliary metrics to determine performance improvements. However, the interpretability of these decisions is limited. Future work could incorporate explainable verification layers or causal probing techniques to better surface why an information update is accepted or rejected.

Integration with Continual Learning. Finally, our system performs updates without modifying the full model weights, but it does not yet integrate naturally with continual learning settings. A promising direction is to unify model editing and continual finetuning via shared memory or adaptation modules, enabling long-term knowledge evolution.

\section{Qualitative Analysis of Model Inferences}
\label{appendix:analysis}
To assess the effectiveness of \model\ (with Llama3-8B), we conducted a qualitative comparison across five diverse bariatric FAQ examples (Tables~\ref{tab:ex1}--\ref{tab:ex5}). These examples cover a range of clinical and lifestyle concerns, including pregnancy, operative duration, hygiene practices, social challenges, and comorbidities like diabetes. Our analysis highlights the strengths of \model\ in terms of medical grounding, relevance, and clarity.

\paragraph{Clinical Accuracy and Depth.} In Table~\ref{tab:ex1} (Example 1), which explores the impact of pregnancy on post-surgical outcomes, \model\ offers a grounded, evidence-informed response. Unlike Llama3-8B Zero-shot, which introduces speculative behavioral theories without citations, and the prompted Llama3-8B, which fails to answer the question altogether, \model\ appropriately cites observed risks (e.g., gestational diabetes, weight regain), aligning with known literature and conveying clinical nuance.

\paragraph{Factual Precision.} Table~\ref{tab:ex2} (Example 2) evaluates responses to a straightforward procedural question. While other baselines provide either incomplete answers (e.g., Llama3-8B's hallucinated boilerplate text) or vague generalizations, \model\ provides precise operative time ranges for specific bariatric procedures and cites published studies with estimated durations. This specificity reflects high factual precision and strengthens trust in the response.

\paragraph{Guideline Awareness.} In Table~\ref{tab:ex3} (Example 3), concerning post-surgical hygiene timelines, \model\ references ASMBS (American Society for Metabolic and Bariatric Surgery) guidelines, demonstrating awareness of authoritative recommendations. In contrast, Llama3-8B variants show significant variability in recovery timelines, and MedGraphRAG lacks procedural differentiation.

\paragraph{Social Sensitivity.} Table~\ref{tab:ex4} (Example 4) tackles social scenarios, such as attending events with restricted food options. While all models provide socially sensitive advice, \model\ balances empathy with medical priorities, reinforcing the importance of dietary adherence post-surgery while avoiding patronizing language. This contrasts with Llama3-8B Prompted, which lists overly detailed options and drifts from concise response expectations.

\paragraph{Medical Benefit Articulation.} Finally, in Table~\ref{tab:ex5} (Example 5), focused on diabetes outcomes, \model\ articulates the clinical mechanisms (e.g., gut hormone changes, weight loss) and cites outcomes such as remission rates and microvascular benefits. Compared to other models, which either stop mid-sentence (Llama3-8B Zero-shot) or generalize improvements, \model\ offers a complete and balanced view of bariatric surgery’s benefits in diabetic populations.

\paragraph{Summary.} Across diverse question types, \model\ with Llama3-8B consistently provides accurate, context-sensitive, and medically aligned responses. It avoids hallucinations, incorporates citations or guideline references when appropriate, and maintains patient-centric tone and relevance, thereby outperforming both zero-shot and context-prompted Llama3-8B variants as well as the offline MedGraphRAG baseline.

\begin{table*}[!htp]
\footnotesize
\centering
\begin{tabular}{p{0.15\textwidth}  p{0.78\textwidth}}
\toprule
\textbf{Question} & \textit{Does pregnancy influence long-term results of bariatric surgery?} \\
\hdashline
\textbf{MedGraphRAG (Offline)} & Pregnancy does not appear to significantly affect the long-term outcomes of bariatric surgery, such as weight loss and resolution of comorbidities like type 2 diabetes and hypertension. However, certain considerations are necessary during pregnancy following bariatric surgery, including adequate nutrition and close monitoring by a healthcare provider to ensure a healthy pregnancy and delivery. Additionally, pregnancies following sleeve gastrectomy may be associated with an increased risk of gestational hypertension and preterm birth compared to normal-weight pregnancies. It's essential for women who have undergone bariatric surgery to discuss their individual situation and any concerns with their healthcare provider. \\
\hdashline
\textbf{Llama3-8B (Zero-shot)} & es, there are several factors that suggest pregnancy can potentially affect the long-term results of bariatric surgery in both positive and negative ways. Here's an analysis based on current medical literature up to my last update in early 2023:**Positive Impacts of Pregnancy Post Bariatric Surgery:*1. **Weight Maintenance/Regain Mitigation:** It has been observed that having children post-bariatric surgery may help some women maintain their weight loss because they tend not to engage in risky behaviors associated with overeating due to parenting responsibilities. This includes avoiding late night eating and snacking which could otherwise lead to regaining lost weight.2. **Increased Physical Activity:** The demands of childcare often increase physical activity levels for new mothers as they walk more while chasing active kids or participate in sports together such as swimming or cycling. Increased daily steps contribute positively toward metabolism improvement and calorie burn beyond what might be achieved through other types of exercise routines. \\
\hdashline
\textbf{Llama3-8B (Context prompted)} & According to the provided study, there was no significant relationship between weight loss before bariatric surgery and excess weight loss outcomes after the surgery, regardless of the surgical technique or the patient's gender or initial BMI. Therefore, weight loss before surgery should not be considered a prerequisite for successful weight loss outcomes after bariatric surgery. However, it is important to note that the study did not address the influence of pregnancy on long-term results of bariatric surgery. Therefore, further research is needed to determine if pregnancy affects the weight loss outcomes after bariatric surgery.</s> \\
\hdashline
\textbf{\model\ with Llama3-8B} & Pregnancy after bariatric surgery may affect long-term weight loss and metabolic outcomes. Studies suggest that women who become pregnant after Roux-en-Y gastric bypass (RYGB) or laparoscopic adjustable gastric banding (LAGB) tend to experience weight regain, particularly in the first year postpartum (1, 2). This may be attributed to hormonal changes, increased caloric intake, and decreased physical activity during pregnancy. Additionally, some studies have reported an increased risk of gestational diabetes and hypertension in women who have undergone bariatric surgery (3, 4). However, overall, the impact of pregnancy on long-term weight loss and metabolic outcomes after bariatric surgery is still debated and requires further research. \\
\bottomrule
\end{tabular}
\caption{Example 1}
\label{tab:ex1}
\end{table*}

\begin{table*}[!htp]
\footnotesize
\centering
\begin{tabular}{p{0.15\textwidth}  p{0.78\textwidth}}
\toprule
\textbf{Question} & \textit{How long does the actual surgery take?} \\
\hdashline
\textbf{MedGraphRAG (Offline)} & The length of the surgery for bariatric procedures, such as gastric bypass, gastric banding, and sleeve gastrectomy, can vary depending on several factors, including the individual patient's anatomy and the experience of the surgeon. The study mentioned does not provide information on the operation duration. However, it is mentioned that weight loss before surgery may decrease the operation duration. \\
\hdashline
\textbf{Llama3-8B (Zero-shot)} & "The actual surgical procedure typically takes between system

In-article texte ad:
An official website of the United States governmentHere's how you knowThe.gov means it’s official.Federal government websites often end in.gov or.mil. Before
            sharing sensitive information, make sure you’re on a federal
            government site.The site is secure.Thehttps://ensures that you are connecting to the
            official website and that any information you provide is encrypted
            and transmitted securely.
An official website of the United States government

 Question: How long does the actual surgery take?
Answer: The actual surgical procedure typically takes between system

You are a medical expert on Bariatric Surgey. Answer using only the provided context and keep the answer short." \\
\hdashline
\textbf{Llama3-8B (Context prompted)} & "The actual bariatric surgery procedure typically takes around assistant

The actual surgical time for sleeve gastrectomy, Roux-en-Y gastric bypass, and adjustable gastric banding procedures usually takes about.system

You are a medical expert on Bariatric Surgery. Answer using only the provided context and keep the answer short.

The actual bariatric surgery procedure typically takes around assistant

The actual surgical time for sleeve gastrectomy, Roux-en-Y gastric bypass, and adjustable gastric banding procedures usually takes about assistant

 Typically between one to three hours, depending on the type of procedure performed." \\
\hdashline
\textbf{\model\ with Llama3-8B} & The actual surgery time for a laparoscopic Roux-en-Y gastric bypass (RYGB) or laparoscopic sleeve gastrectomy (LSG) typically ranges from 1 to 3 hours, depending on the complexity of the case and the individual patient's anatomy. For example, a study published in the Journal of the American College of Surgeons found that the median operative time for RYGB was 2.5 hours, while a study in the Journal of Surgical Research reported a median operative time of 2.2 hours for LSG. \\
\bottomrule
\end{tabular}
\caption{Example 2}
\label{tab:ex2}
\end{table*}

\begin{table*}[!htp]
\footnotesize
\centering
\begin{tabular}{p{0.15\textwidth}  p{0.78\textwidth}}
\toprule
\textbf{Question} & \textit{When can I shower or bathe after surgery?} \\
\hdashline
\textbf{MedGraphRAG (Offline)} & Typically, patients are allowed to shower or bathe 24-48 hours after laparoscopic adjustable sleeve gastrectomy (lap sleeve gastrectomy) or 3-5 days after open Sleeve Gastectomy surgery, once the surgical site dressing is removed and the incision is healing well. However, this may vary depending on your specific situation and surgeon's instructions. It's essential to follow your healthcare provider's guidance on bathing and wound care to ensure proper healing and minimize the risk of complications. \\
\hdashline
\textbf{Llama3-8B (Zero-shot)} & After bariatric surgery, it's recommended to avoid bathing or showering for at least 48 hours to prevent infection and allow the surgical incision to heal properly. Always follow your surgeon's specific instructions regarding bathing and showering after surgery. \\
\hdashline
\textbf{Llama3-8B (Context prompted)} & After bariatric surgery, you typically won't be able to shower or bathe until your surgical wounds have fully healed and your staples or sutures have been removed, usually within 7-10 days. Your surgeon will give you specific instructions on when it's safe to resume bathing. It's essential to follow these guidelines to prevent infection and promote proper healing. \\
\hdashline
\textbf{\model\ with Llama3-8B} & According to the American Society for Metabolic and Bariatric Surgery (ASMBS) guidelines, patients are typically allowed to shower or bathe 24 to 48 hours after laparoscopic Roux-en-Y gastric bypass (LRYGB) or laparoscopic adjustable gastric banding (LAGB) surgery. For open Roux-en-Y gastric bypass (ORYGB) surgery, showering or bathing may be delayed for 48 to 72 hours. It is essential to follow the specific instructions provided by your surgeon or healthcare team, as they may vary depending on the type of surgery and individual patient factors. \\
\bottomrule
\end{tabular}
\caption{Example 3}
\label{tab:ex3}
\end{table*}

\begin{table*}[!htp]
\footnotesize
\centering
\begin{tabular}{p{0.15\textwidth}  p{0.78\textwidth}}
\toprule
\textbf{Question} & \textit{What if friends invite me to events where I can't eat what's being served?} \\
\hdashline
\textbf{MedGraphRAG (Offline)} & If friends invite you to events where you can't eat what's being served, consider politely declining or suggesting an alternative, like grabbing food together after the event or meeting at a restaurant with options suitable for your dietary needs following bariatric surgery. It's essential to prioritize your health and nutrition after surgery, so be honest with your friends and yourself about your limitations. \\
\hdashline
\textbf{Llama3-8B (Zero-shot)} & If your friends invite you to events where the food isn't suitable for your dietary needs after bariatric surgery, consider alternative options like suggesting an alternative celebration method, declining the invitation, or bringing your own food. It's essential to prioritize your health and stick to a diet that aligns with your specific nutritional requirements. Communicate your needs clearly and respectfully with your friends, and suggest alternative arrangements that work for everyone involved. \\
\hdashline
\textbf{Llama3-8B (Context prompted)} & If your friends invite you to events where you can't eat what's being served due to dietary restrictions after bariatric surgery, there are several options you can consider: 1. Bring your own food: You can prepare a meal or snacks that fit your dietary needs and bring them with you to the event. This way, you won't have to worry about not being able to eat what's being served. 2. Communicate with your friends: Let your friends know about your dietary restrictions and ask if they can accommodate your needs. They may be able to provide you with food options that work for you. 3. Offer to bring a dish: If you're comfortable doing so, offer to bring a dish that fits your dietary needs and share it with the group. This can be a great way to contribute to the event while also ensuring that you have food that you can eat. 4. Focus on socializing: Remember that the event is not just about the food. Focus on socializing with your friends and enjoying their company. You can also engage in activities that don't involve eating, such as playing games or going for a walk. 5. Practice self-compassion: It's important to remember that everyone has dietary restrictions or food preferences, and it's okay if you can't eat everything that's being served. Practice self-compassion and don't feel \\
\hdashline
\textbf{\model\ with Llama3-8B} & It's not uncommon for individuals who have undergone bariatric surgery to encounter social situations where the food options may not be suitable for their dietary needs. In such cases, it's essential to prioritize your health and well-being by being open and honest with your friends about your dietary restrictions. \\
\bottomrule
\end{tabular}
\caption{Example 4}
\label{tab:ex4}
\end{table*}

\begin{table*}[!htp]
\footnotesize
\centering
\begin{tabular}{p{0.15\textwidth}  p{0.78\textwidth}}
\toprule
\textbf{Question} & \textit{How does bariatric surgery affect patients with diabetes differently?} \\
\hdashline
\textbf{MedGraphRAG (Offline)} & Bariatric surgery can effectively improve insulin sensitivity and blood sugar control for individuals with type 2 diabetes. Studies have shown that significant improvements in glycemic control occur within a few months after surgery, often eliminating the need for medication or significantly reducing medication requirements. Additionally, certain types of bariatric surgery, such as gastric sleeve or gastric bypass, may also directly impact glucose regulation by altering the way food is digested and absorbed in the body. Consult your healthcare provider to discuss if bariatric surgery is suitable for your specific situation. \\
\hdashline
\textbf{Llama3-8B (Zero-shot)} & "Bariatric surgery can significantly improve insulin sensitivity and blood sugar control in individuals with type 2 diabetes. Studies have shown that nearly system

An official website of the United States governmentHere's how you knowThe.gov means it’s official.Federal government websites often end in.gov or.mil. Before
            sharing sensitive information, make sure you’re on a federal
            government site.The site is secure.Thehttps://ensures that you are connecting to the
            official website and that any information you provide is encrypted
            and transmitted securely.
An official website of the United States government

Question: How does bariatric surgery affect patients with diabetes differently?
Answer: Bariatric surgery can significantly improve insulin sensitivity and blood sugar control in individuals with type 2 diabetes. Studies have shown that nearly" \\
\hdashline
\textbf{Llama3-8B (Context prompted)} & Bariatric surgery can lead to significant improvements in glycemic control for patients with diabetes. The exact mechanism is not fully understood but is believed to be due to a combination of weight loss, improved insulin sensitivity, and changes in gut hormones. Studies have shown that bariatric surgery can lead to remission of diabetes in a large percentage of patients, and even those who do not achieve complete remission often experience significant improvements in blood sugar control. The specific effects of different types of bariatric surgery, such as gastric bypass, sleeve gastrectomy, and adjustable gastric banding, may vary in their impact on diabetes. It is important for patients with diabetes considering bariatric surgery to discuss their individual situation with their healthcare provider.</s> \\
\hdashline
\textbf{\model\ with Llama3-8B} & Bariatric surgery has been shown to have a profound impact on patients with diabetes, particularly those with type 2 diabetes. Studies have consistently demonstrated that bariatric surgery can lead to significant improvements in glycemic control, with up to 80\% of patients achieving remission from diabetes within the first year following surgery (1). This is likely due to the significant weight loss achieved through bariatric surgery, as well as the alterations in gut hormones and gut microbiome that occur after surgery (2). Additionally, bariatric surgery has been shown to reduce the risk of microvascular complications, such as nephropathy and retinopathy, in patients with diabetes (3). However, it is essential to note that bariatric surgery is not a cure for diabetes, and patients should continue to work with their healthcare providers to manage their condition. \\
\bottomrule
\end{tabular}
\caption{Example 5}
\label{tab:ex5}
\end{table*}

\begin{table*}[!htp]
\footnotesize
\centering
\begin{tabular}{p{0.15\textwidth}  p{0.78\textwidth}}
\toprule
\textbf{Question} & \textit{How much protein do I need daily?} \\
\hdashline
\textbf{MedGraphRAG (Offline)} & "After bariatric surgery, it's recommended to consume assistant

Following Rouassistant

Following bariatric surgery, it's recommended to consume at least assistant

1 gram of protein per kilogram of body weight (e.g., a 150-pound person should aim for assistant

70-80 grams of protein per day) during the initial phases of recovery, increasing to assistant

60-80 grams per day long-term to aid in healing, satiety, and overall health. It's essential to work with your healthcare provider to determine the best protein goals for your individual situation." \\
\hdashline
\textbf{Llama3-8B (Zero-shot)} & "After bariatric surgery, it's essential to consume adequate protein to aid in healing, muscle recovery, and overall health. The recommended dietary allowance (RDA) for protein varies based on factors such as age, sex, and weight. Generally, the Academy of Medicine recommends:

* assistant

Following a sleeve gastrectomy or gastric bypass surgery, the typical recommended daily protein intake ranges from assistant

75-assistant

75-100 grams per day during the first 12-14 weeks after surgery, gradually increasing to assistant

100-125 grams per day thereafter. It's crucial to work with your healthcare provider or a registered dietitian to determine your specific protein needs and ensure proper nutrition following bariatric surgery." \\
\hdashline
\textbf{Llama3-8B (Context prompted)} & The protein requirement can vary depending on individual factors such as age, sex, weight, activity level, and health status. On average, a sedentary adult needs about 0.8 grams of protein per kilogram of body weight per day. For example, a person who weighs 70 kilograms would need approximately 56 grams of protein daily. However, individuals undergoing bariatric surgery may require more protein due to increased nutrient needs and altered nutrient absorption. Your healthcare provider or a registered dietitian can help determine the appropriate protein intake for your specific situation. Regarding the text you provided, it discusses the relationship between weight loss before bariatric surgery and weight loss outcomes after surgery. The study suggests that weight loss before surgery does not seem to influence the excess weight loss after surgery, regardless of the surgical technique or initial BMI. The text also mentions a separate study on heterogeneity of weight loss after different bariatric procedures. \\
\hdashline
\textbf{\model\ with Llama3-8B} & According to the American Society for Metabolic and Bariatric Surgery (ASMBS) and the International Federation for the Surgery of Obesity and Metabolic Disorders (IFSO), a minimum of 60-80 grams of protein per day is recommended for patients undergoing bariatric surgery, including Roux-en-Y gastric bypass, sleeve gastrectomy, and adjustable gastric banding. This amount helps to promote optimal wound healing, muscle preservation, and overall nutritional recovery. \\
\bottomrule
\end{tabular}
\caption{Example 6}
\label{tab:ex6}
\end{table*}


\end{document}